\documentclass[sn-mathphys-num]{sn-jnl}

\usepackage{graphicx}%
\usepackage{multirow}%
\usepackage{amsmath,amssymb,amsfonts}%
\usepackage{amsthm}%
\usepackage{mathrsfs}%
\usepackage[title]{appendix}%
\usepackage{xcolor}%
\usepackage{textcomp}%
\usepackage{manyfoot}%
\usepackage{booktabs}%
\usepackage{algorithm}%
\usepackage{algorithmicx}%
\usepackage{algpseudocode}%
\usepackage{listings}%

\theoremstyle{thmstyleone}%
\theoremstyle{thmstyletwo}%

\theoremstyle{thmstylethree}%

\begin{document}

\title[Article Title]{Robot Excavation and Manipulation of Geometrically Cohesive Granular Media}

\author[1]{\fnm{Laura} \sur{Treers}}\email{laura.treers@uvm.edu}
\equalcont{These authors contributed equally to this work.}

\author[2]{\fnm{Daniel} \sur{Soto}}\email{dsoto7@gatech.edu}
\equalcont{These authors contributed equally to this work.}

\author[3]{\fnm{Joonha} \sur{Hwang}}\email{jhwang321@gatech.edu}

\author[4]{\fnm{Michael A. D.} \sur{Goodisman}}\email{michael.goodisman@biology.gatech.edu}

\author*[2]{\fnm{Daniel I.} \sur{Goldman}}\email{daniel.goldman@physics.gatech.edu}

\affil[1]{\orgdiv{Department of Mechanical Engineering}, \orgname{University of Vermont}, \orgaddress{\street{33 Colchester Ave.}, \city{Burlington}, \postcode{05405}, \state{Vermont}, \country{USA}}}

\affil[2]{\orgdiv{School of Physics}, \orgname{Georgia Institute of Technology}, \orgaddress{\street{837 State St.}, \city{Atlanta}, \postcode{30332}, \state{Georgia}, \country{USA}}}

\affil[3]{\orgdiv{George W. Woodruff School of Mechanical Engineering}, \orgname{Georgia Institute of Technology}, \orgaddress{\street{801 Ferst Dr NW}, \city{Atlanta}, \postcode{30318}, \state{Georgia}, \country{USA}}}

\affil[4]{\orgdiv{School of Biological Sciences}, \orgname{Georgia Institute of Technology}, \orgaddress{\street{310 Ferst Dr. NW}, \city{Atlanta}, \postcode{30332}, \state{Georgia}, \country{USA}}}


\abstract{Construction throughout history typically assumes that its blueprints and building blocks are pre-determined. However, recent work suggests that alternative approaches can enable new paradigms for structure formation. Aleatory architectures, or those which rely on the properties of their granular building blocks rather than pre-planned design or computation, have thus far relied on human intervention for their creation. We imagine that robotic swarms could be valuable to create such aleatory structures by manipulating and forming structures from entangled granular materials. To discover principles by which robotic systems can effectively manipulate soft matter, we develop a robophysical model for interaction with geometrically cohesive granular media composed of u-shape particles. This robotic platform uses environmental signals to autonomously coordinate excavation, transport, and deposition of material. We test the effect of substrate initial conditions by characterizing robot performance in two different material compaction states and observe as much as a 75\% change in transported mass depending on initial substrate compressive loading.  These discrepancies suggest the functional role that material properties such as packing and cohesion/entanglement play in excavation and construction. To better understand these material properties, we develop an apparatus for tensile testing of the geometrically cohesive substrates, which reveals how entangled material strength responds strongly to initial compressive loading. These results explain the variation observed in robotic performance and point to future directions for better understanding robotic interaction mechanics with entangled materials. }


\maketitle

\section{Introduction}\label{sec1}

\subsection{Background}

The growing need to improve construction safety and sustainability has led to increasing interest in developing robotic teams for collective construction \cite{petersen_review_2019}. Such modern robot teams have many potential applications, from the construction industry to disaster response to space exploration. Existing experimental mobile robotic systems have largely drawn from concepts more typical in conventional construction, utilizing discrete building elements such as rigid bricks, struts, or blocks \cite{werfel_designing_2014, seo_assembly_2013}. Some recent work has explored the use of amorphous materials such as foams \cite{napp_distributed_2014,saboia_autonomous_2019} toothpicks with applied adhesive \cite{napp_materials_2012}, or on-site materials such as pebbles \cite{saboia_da_silva_autonomous_2019,furrer_autonomous_2017,thangavelu_dry_2018}. Some researchers have also developed tools for robotic manipulation of granular media through perception \cite{matl_interactive_2021} or learning \cite{pmlr-v78-schenck17a, tuomainen_manipulation_2022}. However, these explorations have largely focused on simulated settings and/or fixed robotic arms. Strategies for successful cohesive material manipulation remain poorly understood. 

Recent studies have demonstrated that embracing disorder in design can enable new functionalities in structures. This design approach, termed aleatory architecture, suggests that structural stability can emerge via particle jamming, rather than through precise placement of these particles \cite{keller2016aleatory, murphy2017aleatory}. In turn, aleatory structures can be ``self-healing," and can be built in environments where utilizing pre-determined blueprints and construction materials is impossible. Architects and designers have explored this new design paradigm in recent years, and have demonstrated that a variety of particle shapes and sizes can be utilized for creating rigid structures with little pre-planning \cite{dierichs_towards_2016, dierichs2021designing}. These types of structures can be formed via confinement of particles in external membranes, or internal struts. However, varying particle geometry represents another means of generating inter-particle cohesion without relying on membranes, or other interstitial fluid or adhesives \cite{zhao2016packings}. 

\begin{figure}[!h]
    \centering
    \includegraphics[scale=0.5]{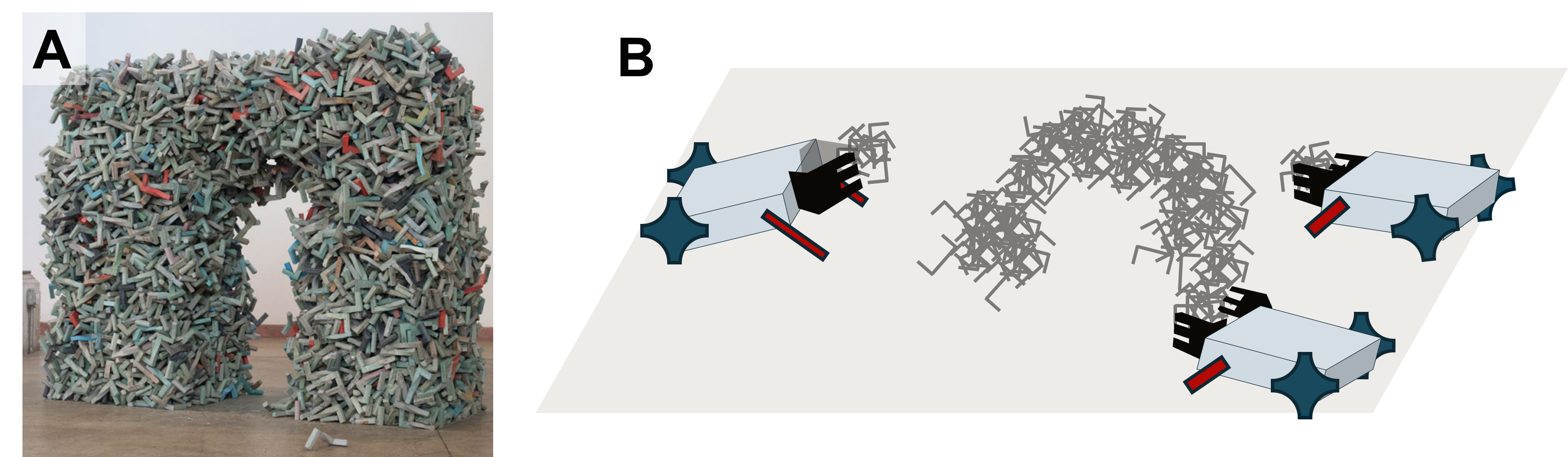}
    \caption{\textbf{ \textbf{Aleatory architectures and robotic construction. } (A)} Example of an aleatory structure realized through concave, geometrically entangled particles. Adapted from \cite{murphy2017aleatory}. \textbf{(B)} Concept for robot teams building aleatory structures. }
    \label{fig:intro_aleatory} 
\end{figure}

Despite the promise of these new methods, they have largely been explored in the context of the final structure \cite{keller2016aleatory}. Aleatory structures that have been explored thus far have relied on pouring or forming cohesive particles into molds \cite{murphy2017aleatory} (Figure \ref{fig:intro_aleatory}A). 
Instead, we hypothesize that such aleatory systems are well-suited for collective robotic construction: in particular, geometrically entangled materials, such as that shown in Figure \ref{fig:intro_aleatory}A, are able to provide strong cohesive forces without the requirement for adhesives or interstitial fluid \cite{savoie_amorphous_2023, savoie_smarticles_2019, gravish_entangled_2012, gravish_entangled_2016}, making them suitable for repeatable manipulation and transport. Additionally, because geometrically entangled substrates do not rely on interstitial fluid, which can be difficult to prepare repeatably, their material properties can be well-characterized in the laboratory setting. 

However, relatively few works until recent decades have explored the rheology of these complex substrates. Prior work has shown that longer bulks of entangled materials are weaker than shorter ones \cite{franklin_extensional_2014}, and that bulk strength is sensitively dependent on particle geometric properties \cite{gravish_entangled_2012, pezeshki_tunable_2024}.  More recent work has also begun to explore their extensional rheology, demonstrating stick-slip behavior during tension caused by local yielding and rearrangement of particle entanglements \cite{franklin_extensional_2014, pezeshki_tunable_2024}. However, many questions remain unanswered with regard to the behavior of these materials, including their response to external loading. Some recent work has demonstrated a stress-induced anisotropy \cite{karapiperis_stress_2022}, as well as compression and shear-driven jamming behavior \cite{marschall_compression-_2015}; however, much of these explorations of stress-dependent material behavior remain in simulated environments. Better understanding these non-traditional material behaviors in an experimental setting will be important for developing strategies for manipulation and transport of these substrates. 

The manipulation of such materials represents a particularly challenging task, as it requires tearing or separating some subset of particles from the larger bulk, and subsequent (or simultaneous) re-forming or consolidation of that subset of particles for transport and/or deposition. Despite the rich nature of these interaction mechanics, the principles by which cohesive materials are manipulated, and by which they can be then utilized to form structures, remain unknown. For example, which types of manipulation strategies result in successful separation of material? Similarly, what are effective methods for re-strengthening cohesive materials once separated? And lastly, how should agents decide if and where to begin excavating materials? 

In this work, we seek to begin to discover principles for robust manipulation of geometrically cohesive materials. We do so by developing a robophysical model \cite{aguilar2016review} capable of autonomously excavating and transporting geometrically cohesive substrate. We study trends in excavation performance during autonomous operation for various material preparation states. We then seek to understand the robotic performance trends and test strategies for substrate manipulation via material tensile testing, and demonstrate a strong correlation between external loads and subsequent tensile strength in entangled materials. The results from this robophysical study have implications for the future design of individual robot manipulation strategies, as well as for the future deployment of robotic teams.

\begin{figure}[!h]
    \centering
    \includegraphics[scale=0.8]{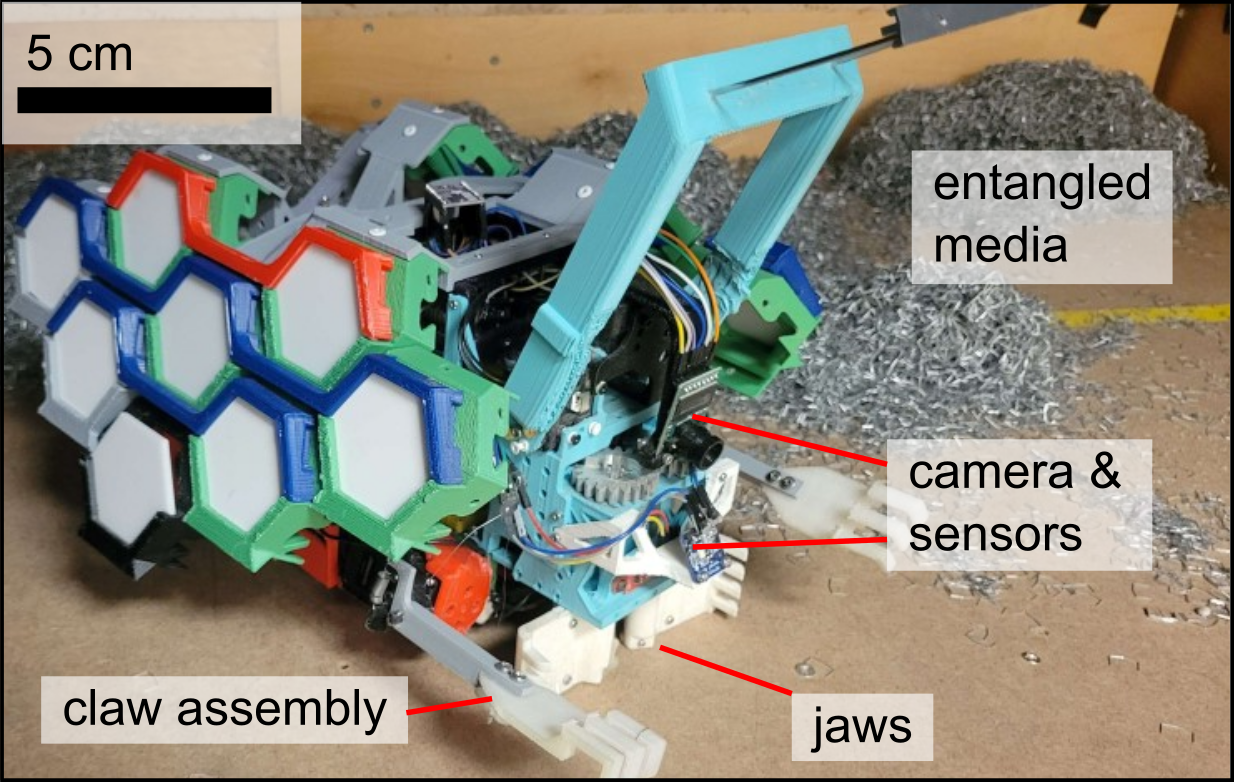}
    \caption{\textbf{Image of robophysical model used in this work}. We use the model to study the manipulation and transport of geometrically cohesive granular material, which can be used to inform future engineering design.} 
    \label{fig:intro}
\end{figure}

\section{Robophysical Model for Interaction with Cohesive Media}\label{substrate}

\subsection{Design of Robophysical Model and Testing Environment}
\label{sec:Robophysical model mat methods}
Our robotic agent for manipulating cohesive granular media is a 3D-printed modified minimal locomotor design (length and width = 32 cm, height = 18 cm) (Figure \ref{fig:intro}) with 5 Dynamixel AX12A motors: two 2-DoF front limbs equipped with 3D-printed nylon ``claws" and one motor driving a pair of rear 3D-printed whegs \cite{quinn2002improved} that are equipped with rubber treads for traction. An external shell protects internal electronics from particle ingress. The robot is battery powered and utilizes a suite of sensors for navigation and material sensing, which are described in Appendix \ref{secA1} .

We use U-shaped particles (office staples) as geometrically entangled substrate in this work \cite{gravish_entangled_2012}. Each particle has a length of 12mm along its longest axis and two perpendicular segments of length 6mm.  We detached 500,000 staples from one another using acetone to dissolve the binding adhesive.  The resulting material is capable of forming amorphous 3D structures without the need for chemical or capillary cohesion.

The robot operates within a constructed ``arena" and uses lights to coordinate movement between an ``excavation zone" and ``deposition zone". The details of this arena setup are described in Appendix \ref{secA2}. The robotic agent is equipped with a finite state machine (FSM) based on the testing environment described above (a full FSM is provided in Appendix \ref{secA3}). The FSM embodies the different behaviors needed for the robotic agent to excavate material, transport it to the deposition site, and finally return to the excavation site to repeat the process. The transitions between behaviors are governed by environmental signals which are detected via onboard sensors. The deposition process involves a search algorithm to find existing piles on which to deposit pellets, with the goal of creating emergent structures from cohesive material; this search process is also described in the Appendix. However, in this work, we focus specifically on the mechanics of interaction required to tear and separate material during the excavation process. 

\subsection{Results \& Discussion}
\subsubsection{Excavation and Material Manipulation Strategy}

\begin{figure}[!h]
    \centering
    \includegraphics[scale=0.65]{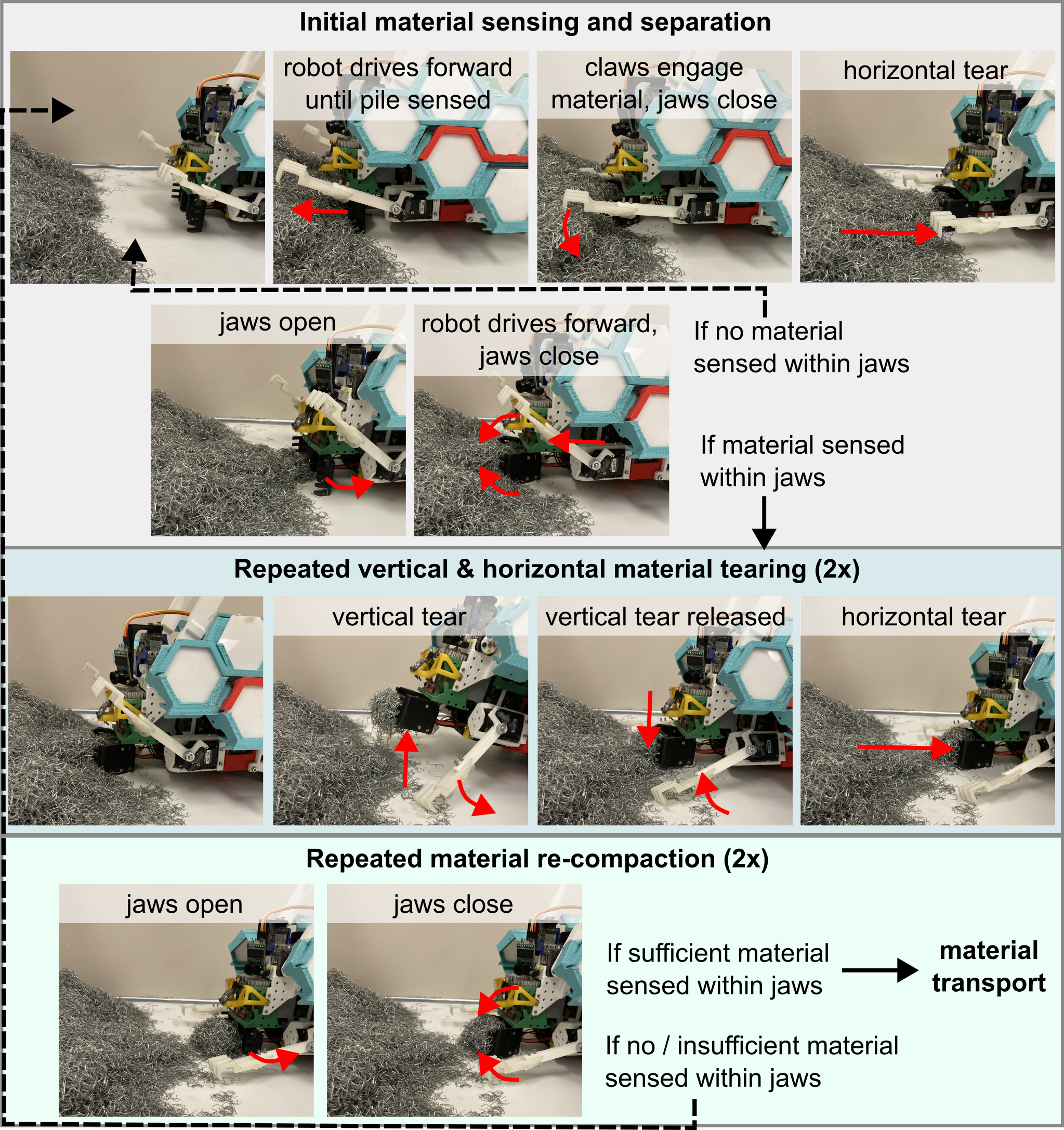}
    \caption{\textbf{Illustrations of robot excavation strategies. } Depiction of excavation sequence, including grasping and tearing geometrically entangled media.}
    \label{fig:robot_excavation}
\end{figure}

The robot manipulates geometrically cohesive material using a set of interlocking jaws, which allow the robot to penetrate and engage small regions of a staple mound (Figure \ref{fig:robot_excavation}). These jaws are 3D printed and are run by a single Dynamixel AX12A motor. The process of removing ``pellets" of material from a larger mound requires repeated, large stressing events, more than what the jaws alone could provide. In order to successfully separate pellets from the larger bulk, we designed an excavation procedure, illustrated in Figure \ref{fig:robot_excavation}. We discover that a sequence of initial material sensing and horizontal separation, followed by repeated vertical and horizontal tearing (using the limbs and rear whegs, respectively), enables material separation. By engaging the limbs during the excavation procedure, the robot effectively weakens the connection between the pellet within the mouth and the rest of the mound, thus allowing the robot to remove a subset of material and transport it elsewhere. 

Lastly, we discover that separation alone through tearing is not sufficient for material transport, because the separated material is often weak and less cohesive, leading to dropping of individual particles during transport. We introduce a final step to the excavation procedure which repeatedly re-compacts the separated material using the jaws, creating a cohesive pellet that can then be transported.  The number of cycles needed to remove the pellet from the mound varies with the size of the pellet, the material packing state, and the local topology of the mound. Given that these conditions are highly variable and can change over time, the robot does not perform a fixed number of excavation procedures. Instead, the robot runs this protocol until the jaws are unable to fully close after the robot has backed away from the mound. This condition is measured using the Dynamixel internal encoders, which indicate whether the jaw angle is greater than a threshold of 
$\approx$30$^o$. This state signifies that material is in the jaws and that the robot has successfully removed a ``pellet" from the mound.

\subsubsection{Evaluation of Performance}
We evaluate the performance of the robot excavating and transporting material with three metrics: average cycle time, excavation success rate, and total material transported. The first two metrics are computed through manual video parsing. Within each cycle, excavation success is determined through the oblique camera: if the agent is able to break off material from the excavation pile and retain at least 1/4 of the jaws' capacity ($\approx$3 cm$^3$) by the end of its turning maneuver, it is marked as an excavation success. Once a trial is complete, we gather the material at the deposition site and weigh it, providing a final performance metric. These methods serve to quantify robot performance across trials and help characterize the challenges associated with cohesive material manipulation.

\begin{figure}[!h]
    \centering
    \includegraphics[scale=1]{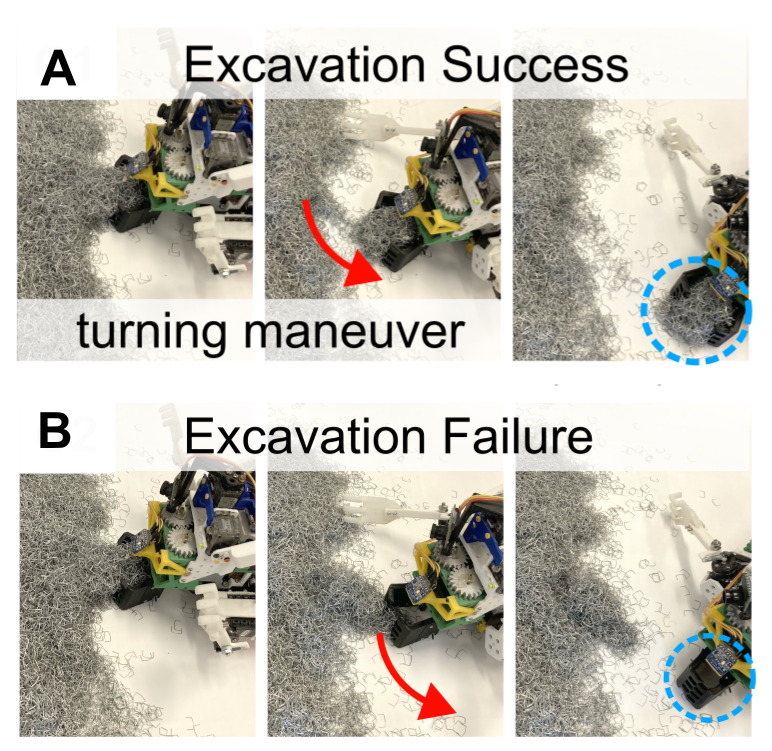}
    \caption{ \textbf{Depiction of robot excavation performance. } \textbf{(A)} Robot excavation success, in which a cohesive ball of substrate is formed and transported. \textbf{(B)} Excavation failure, in which the substrate excavated is either dropped or not separated from the bulk. }
    \label{fig:results_success_failure}
\end{figure}

\subsubsection{Excavation Performance}

\begin{figure}[!h]
    \centering
    \includegraphics[scale=0.75]{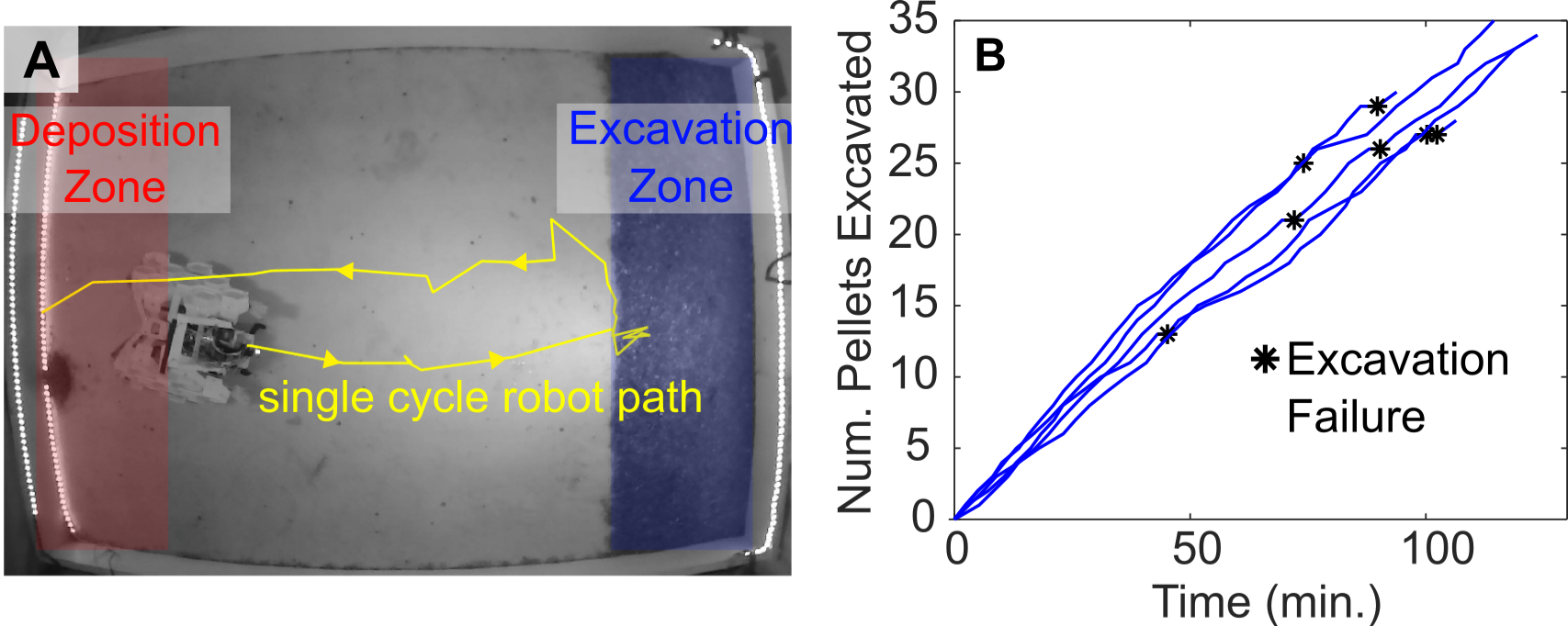}
    \caption{ \textbf{Robotic excavation and deposition cycles. } \textbf{(A)} Depiction of robot path in its testing environment, during a cycle of excavation and deposition. \textbf{(B)} Number of pellets excavated over time for 5 trials. Asterisks indicate excavation failures.  }
    \label{fig:results_init}
\end{figure}

We seek to characterize this robotic agent's performance and in turn understand the strategies which result in successful material manipulation. In a first set of experiments, we seek to replicate trails with repeatable material characteristics to identify both successful material transport as well as failure modes. We first prepare the arena such that piles of staples are shaken from ~0.3m above the arena floor, such that individual staples or very small groups fall within a fixed rectangular region (1.2 x 0.3m) adjacent to the wall. Over the course of two hours, the robot proceeds through cycles of excavation and deposition (Figure \ref{fig:results_init}A). As shown in Figure \ref{fig:results_init}B, the number of pellets excavated grows approximately linearly in all five trials tested. However, particularly late in the trial, we observe the emergence of a failure mode in the excavation process: when the robot has completed multiple cycles of tearing, backs away from the pile, and still senses material within its jaws, the algorithm infers that material has been separated and the robot switches to transport. However, depending on the relative strength of the substrate, the material within the jaws may be ``stretched" but still attached to the main bulk via strong tensile forces, and thus the robot is unable to separate the material within its jaws for transport. Thus, eventually the material slips out of the jaws during the attempts to turn, which we denote an excavation failure, as shown in Figure \ref{fig:results_success_failure}B (denoted with asterisks in Figure \ref{fig:results_init}B). 

Since we observe most excavation failures occurring late in trial, we hypothesize that the repeated cycles of excavation may disturb and compress the substrate over time, leading to a relative strengthening via entanglement of particles. Because the robot is only able to sense the amount of material it has grasped, and not the force required to tear away that material, we hypothesize excavation failures are more likely to occur when the substrate has higher tensile strength. These results suggest the possibility of a compression-dependent strengthening of entangled substrate, which we explore via controlled variation in material properties. 

\subsubsection{Effect of material properties on robot behavior}
We seek to test our prior hypotheses and further investigate how material properties impact the excavation  process. Thus, we conduct two sets of experiments in which we prepare the material at the excavation site in one of two methods: ``scattering" and ``pushing."  

In the ``scattering" preparation mode, identical to the prior section, piles of staples are shaken such that individual particles fall within the fixed region adjacent to the arena wall. In the ``pushing" mode, the same number of staples are instead scattered across the entire arena area, and then are pushed to be contained within the same rectangular region. The volume fractions of portions of staples from each of these preparations were $0.072 \pm 0.005$ and $0.117 \pm 0.002$ for pushing and scattering, respectively. We hypothesize that the disturbances which occur during pushing may alter the media strength, and thus these two preparation modes probe the sensitive dependence of manipulation performance on substrate characteristics.

\begin{figure}[!h]
    \centering
    \includegraphics[scale=0.7]{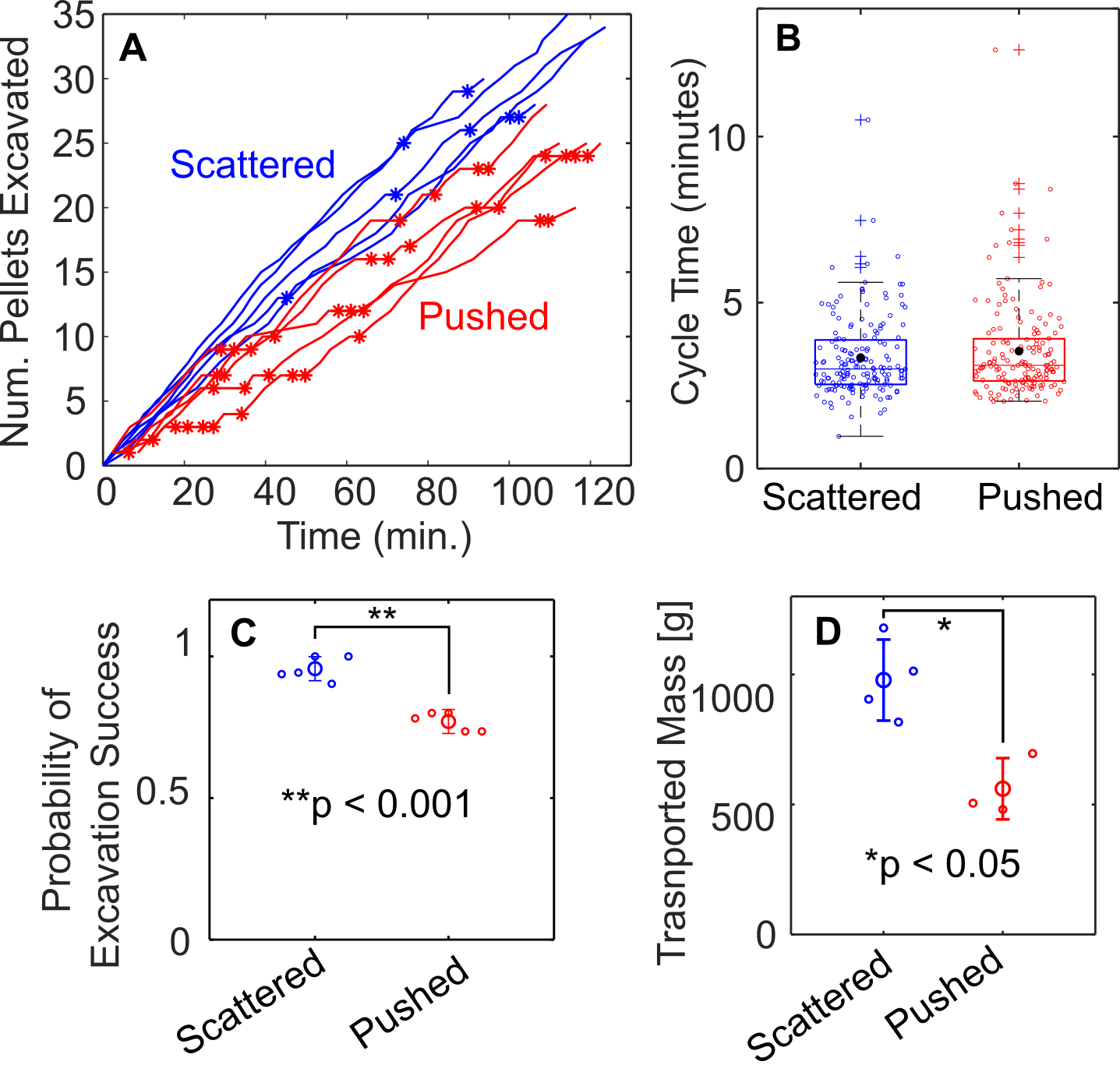}
    \caption{ \textbf{Pellet excavation performance for two material preparation states. } \textbf{(A)} Number of pellets excavated over time for 5 trials of each tested material condition. Asterisks indicate excavation failures. \textbf{(B)} Cycle times for all trials compared between scattered and pushed material states. Plus signs indicate outliers. \textbf{(C)} Comparison in mean excavation success over two hours between scattered and pushed initial material conditions. \textbf{(D)} Comparison of total transported mass of material at the end of trials for both scattered and pushed states.}
    \label{fig:results2}
\end{figure}

After 5 trials of approximately 2 hours for both the scattered and pushed material states, we analyze the proportion of successful excavation over 2 hours as a function of the initial material states. Due to the agent's limited sensing capabilities and uncertainty in media state, it is not guaranteed to remove material each time it exits the excavation area. Thus, we denote whether material is successfully removed at every cycle, and observe that more pellets are excavated successfully over time in the scattered condition (Figure \ref{fig:results2}A). In turn, we find that the excavation probability is dependent on initial material preparation (Figure \ref{fig:results2}C). In the scattered condition, the robot successfully removes material with a probability of $0.96 \pm 0.04$ whereas in the pushed initial condition, this likelihood decreases to $0.77 \pm 0.03$. We observe statistical significance in excavation success between conditions with p = 0.00054. Despite these differences in excavation probabilities, we see little statistical difference between the distributions of cycle times for each initial material state (Figure \ref{fig:results2}B), with scattered and pushed cycles taking $3.3 \pm 1.2$ and $3.5 \pm 1.5$ minutes, respectively. 

While we observe a 75\% increase in mass transported from the pushed to the scattered condition (Figure \ref{fig:results2}D), these ratios in performance are not reflected by the difference in excavation success. Instead, we observe only a 25\% increase in excavation success rate from the pushed to scattered condition. This discrepancy might be explained by differences in transported pellet size and mass, or differences in relative density of the transported material between these two states. 

In summary, these results demonstrate that on a robophysical model for cohesive substrate interaction, manipulation performance is highly sensitive to material properties. Furthermore, we hypothesize that the significant discrepancies in performance may be attributable to variation in material strength between the two preparation states. In the ``pushed" preparation mode, it is possible that compression or other disturbance to the media leads to different material behavior than in the ``scattered" case, thus leading to wide variation in robotic performance. We then seek to probe the dependence of entangled media behavior on prior loading via tensile experiments. 

\section{Entangled Substrate Characterization}\label{substrate}

\subsection{Substrate Tensile Test Apparatus}
\label{sec:tensile testing1}

\begin{figure}[!h]
    \centering
    \includegraphics[scale=0.45]{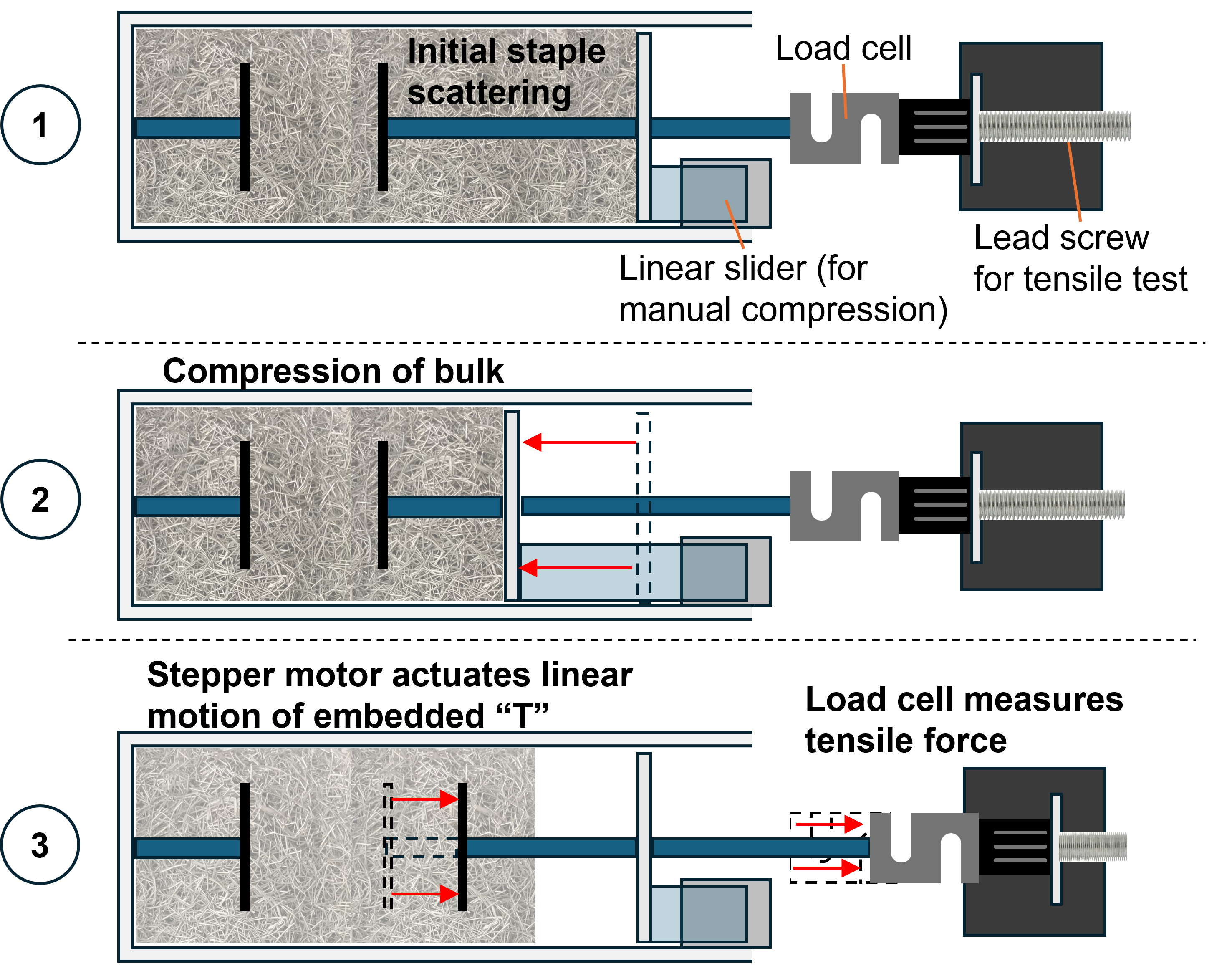}
    \caption{\textbf{Diagram of staple testing apparatus. } \textbf{(1)} Staples are initially scattered in fixed area defined the position of a linear slider. \textbf{(2)} Linear slider is used to manually compress the staple sample, and the slider is subsequently returned to its initial position so that the staple bulk can later be expanded. \textbf{(3)} A stepper motor/lead screw assembly is used to actuate linear motion of the rightmost embedded T shape. The load cell records tensile forces.}
    \label{fig:staple_pit}
\end{figure}

To test the effects of initial loading on the strength of geometrically entangled materials, we construct a materials testing apparatus (see Fig. \ref{fig:staple_pit}). The setup enables both controlled compressive loading of a bulk of staples, as well as subsequent tensile stress on this bulk. A single axis load cell (CALT DYLY 30Kg S Beam Load Cell) measures tensile forces during pulling, and linear motion is generated via a lead screw and stepper motor (Nema 23). We measure stress-strain relationships of a bulk of staples (1 kg of substrate distributed across a 26x10.1x11.7cm$^3$ open top box), for different initial conditions. Staples are originally scattered in a subset of a 26cm long box and are manually compressed various distances (0, 1.25, 2.5, and 5cm) to a final staple sample length of 18.9cm. T-shapes, which are embedded in the box during staple scattering, are pulled apart with the lead screw assembly for a distance of 6.4 cm, and forces are recorded with the load cell and an amplifier chip (HX711) through serial to an Arduino Uno. We run three tests for each compression value, manually resetting and compressing the staples to the desired distance before each tensile test. Note that in this apparatus the staple bulk is not pulled to full separation -- in future experiments, measuring tensile forces during full extension may reveal further properties of these entangled substrates.

\subsection{Substrate Testing Results}
\label{sec:tensile testing2}

\begin{figure}[!h]
    \centering
    \includegraphics[scale=0.75]{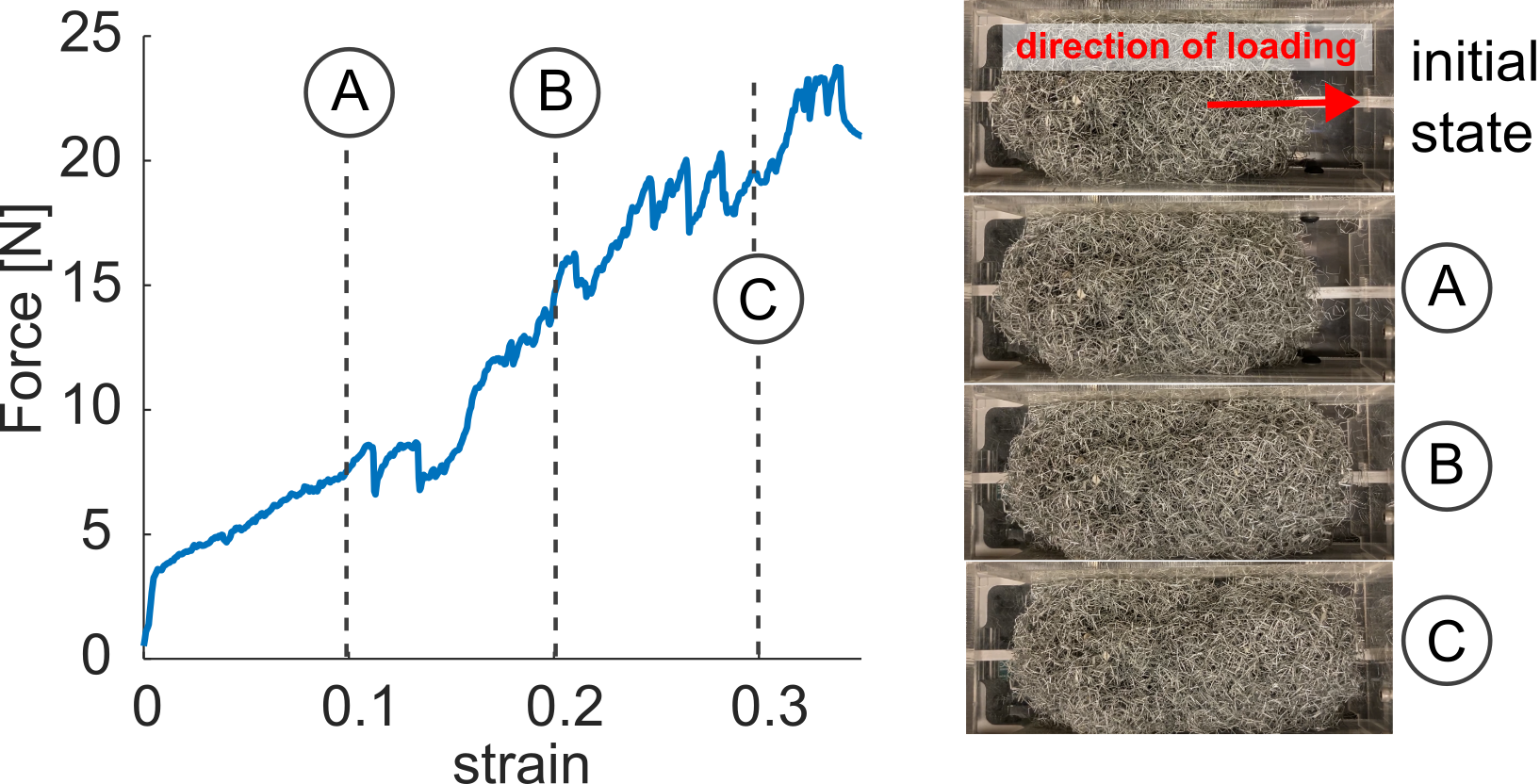}
    \caption{\textbf{Single trial tensile testing data. }\textbf{(Left)} Sample force vs. strain relationship for a single trial of a staple bulk with 10 \% initial compression.  \textbf{(Right)} The corresponding photos of the staple bulk at different strain values. Direction of loading during tensile force vs. strain experiments is indicated. }
    \label{fig:stress_curve}
\end{figure}

The measured force vs. strain relationship for small amounts of compression (a trial at 10\% compression is shown in Fig. \ref{fig:stress_curve}) demonstrates a broadly linearly increasing trend, but is non-monotonic. This trend also includes local yield events which lead to momentary decreases in tensile force, which we hypothesize corresponds to bending or breaking of entangled contacts between individual or groups of staples. These ``local yielding" events occurring during bulk tensile forces have been observed in prior observations of entangled media behavior \cite{franklin_extensional_2014}.

\begin{figure}[!h]
    \centering
    \includegraphics[scale=0.7]{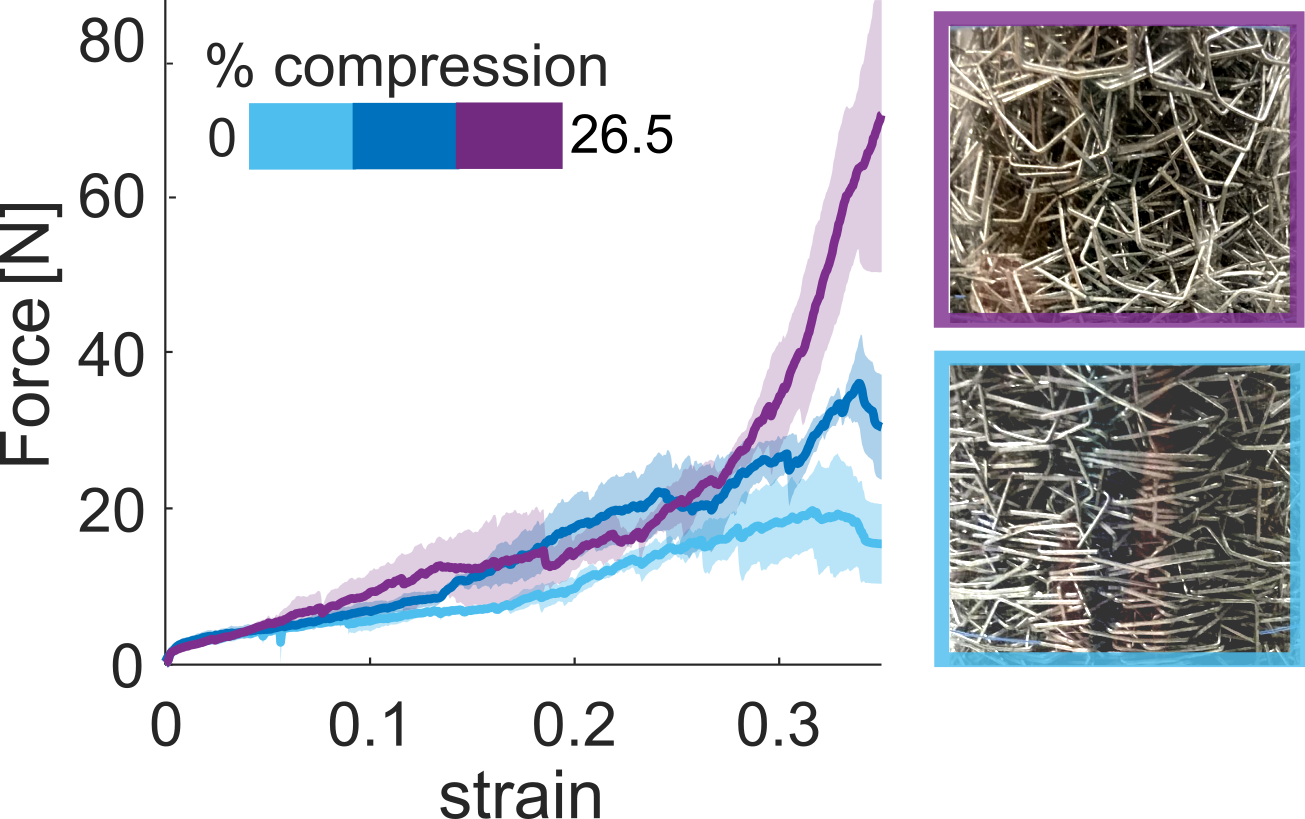}
    \caption{ \textbf{Force-strain relations for staple samples with different initial compressions.} Means and standard deviations over three trials are represented with solid lines and shaded regions, respectively. \textbf{(Sidebar)} Two photos of scattered (blue) and pushed (purple) material samples show that previously compressed staples are less likely to be geometrically aligned, and thus, are more entangled. }
    \label{fig:staple_testing1}
\end{figure}

\begin{figure}[!h]
    \centering
    \includegraphics[scale=0.7]{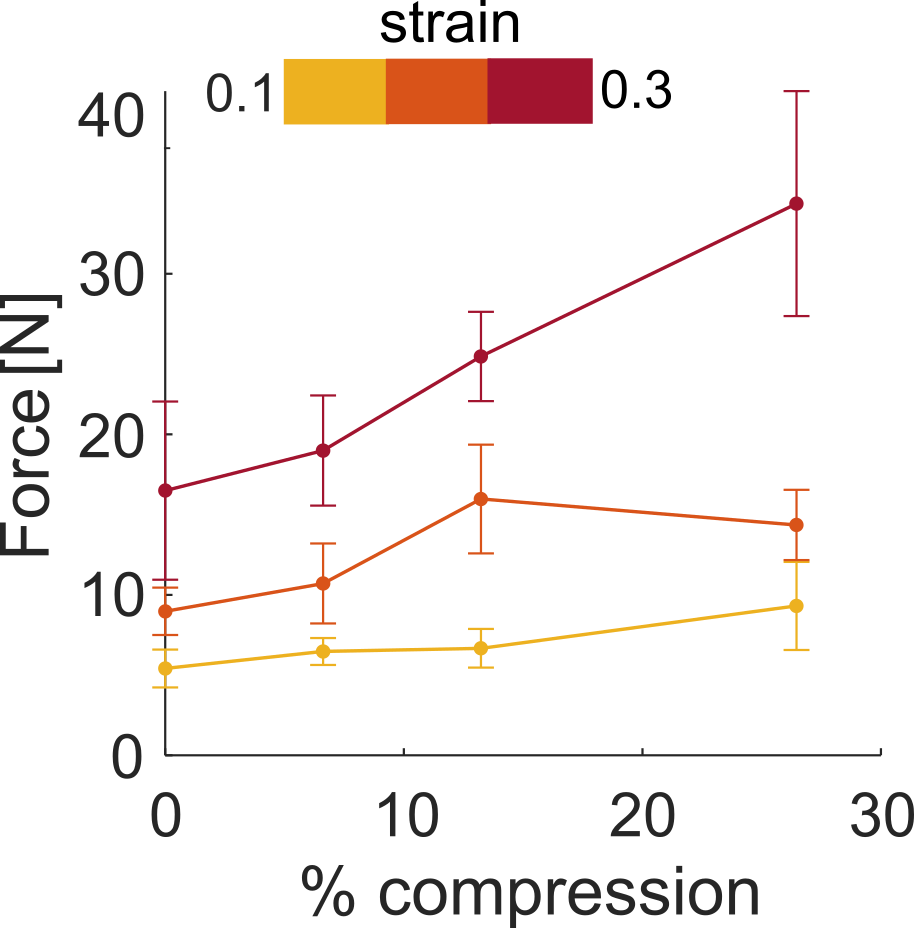}
    \caption{\textbf{Effect of initial compression on tensile force}. Tensile force vs. various initial compression, for three different strain amounts. Error bars indicate standard deviation over three trials. The effect of initial compression on tensile strength is most evident at higher strain values. }
    \label{fig:staple_testing2}
\end{figure}

As shown in Figure \ref{fig:staple_testing1}, for higher values of initial compression, we observe a stress-strain relation for the staple bulk which has a slow initial rise that sharply increases at higher strain values. We generally observe an increase in tensile force for greater initial compression values; however, as demonstrated in Figure \ref{fig:staple_testing2}, this increase is more drastic at higher values of strain. In other words, the role of initial compression on material strength is most apparent when the substrate has begun to separate while in tension. For example, at 30\% strain, the 26\%-compressed staple bulk on average exerted 34.4 $\pm$ 7N of tensile force, nearly double the 16.5 $\pm$ 5.5N registered by the uncompressed sample. In contrast, at 10\% strain, the 26\%-compressed staple bulk exerted 9.3 $\pm$ 2.7N of tensile force, on par with the 5.4 $\pm$ 1.2N from the uncompressed sample. 

These results demonstrate the complex rheological properties of entangled substrates, as they respond differently depending on prior loading. In the robophysical experiments, the ``pushed" material state has some initial compression (which likely varies throughout the substrate), while the ``scattered" case has little to no initial compression. We posit that the behavior observed in our tensile tests points to the consistency observed in excavation of pellets during the ``scattered" trials, in that tensile forces remain low across strain values (throughout the separation process). In contrast, if the ``pushed" material was compressed during preparation, our results suggest that not only will the excavation process require much more tearing force, but there will be greater variability in these forces, especially at high strain or once the material has been initially ``stretched." As a result, our robotic platform may have been capable of initial material separation, but due to the entanglement elicited by initial compression, very large tensile forces were still present in the bulk which led to occasional excavation failures in the ``pushed" scenarios.   We believe the frequent excavation failures observed during these trials may be attributed to this unusual compression-dependent strengthening of entangled materials. 

\section{Conclusion \& Future Work} \label{conclusion}

In this paper, we present a robophysical model that is capable of manipulating and transporting geometrically entangled media. This agent represents one of the first mobile platforms to successfully tear and manipulate highly cohesive substrate. We demonstrate that the robot can operate autonomously over several hours, basing decisions only on local environmental inputs. We also study how robot performance is affected by the properties of the cohesive substrate, and show that preparation of entangled media can significantly affect overall excavation performance. We identify several important areas of continued research, including the rheology of geometric entanglement and structure formation with soft matter.

As evidenced by the discrepancies in robot performance observed, better understanding the complex behavior of cohesive and entangled substrates will be paramount for future work. We observe several unexpected behaviors in this work, including ``pushing" of scattered U-shaped particles resulting in a \emph{decrease} in volume fraction by 39\% relative to the baseline ``scattered" state. We also observe that previous compression of a bulk of entangled media can result in up to a 108\% increase in force required to tear the material. These counterintuitive results highlight that, in addition to packing and entanglement, other non-traditional material parameters such as particle alignment may be necessary to characterize the behavior of entangled substrates. In future work, we seek to leverage rheological understanding to better tune both excavation sequences and jaw geometry to material properties. 

In future work, we posit that similar platforms can be scaled up to serve as model robophysical systems for construction with cohesive media, and provide guidance for future robot teams.  Future studies may also seek to develop principles of structure formation with amorphous material. Our study provides preliminary evidence that mounds of material may emerge over time despite imperfections in localization to previous deposits.  Future work may seek to leverage this characteristic of entangled media to enable the formation of aleatory architectures. Such construction tasks may also be enabled via multi-agent robotic teams, and work should  explore the role of collective interaction with cohesive material.

Robotic teams capable of achieving these complex construction tasks would be of value in a variety of remote construction settings both on earth and in space. For example, recent NASA missions such as the Artemis project have targeted constructing bases on the moon and other planets \cite{von_ehrenfried_artemis_2020,khoshnevis_automated_2004}. However, transporting external material to these locations is often too costly or prohibitive. Small mobile robots capable of building structures from lunar regolith or other found material could alleviate problems associated with material transport \cite{rossi_computational_2024,mueller_construction_2017}. Similarly, in remote or undeveloped locations on earth, robotic agents capable of constructing from resources in their local environments could reduce costs associated with construction equipment and shipping \cite{gerling_robotics_2016,pradhananga_identifying_2021} or assist in disaster response through rubble removal and reorganization \cite{gregory_application_2016}. 

Lastly, future work could use robophysical platforms as tools to understand principles of collective construction observed in many biological species. Many organisms rely on the construction of nests for survival \cite{hansell_animal_2005}; for example, social insects build nests from cohesive substrates \cite{tschinkel_fire_2013, holldobler_ants_1990, monaenkova_behavioral_2015}, beavers build dams using logs and branches \cite{larsen_dam_2021}, and birds build nests using various elongate particles \cite{andrade-silva_cohesion_2021, weiner_mechanics_2020, bhosale_micromechanical_2022}. The aforementioned behaviors all require manipulation of particles which demonstrate cohesive properties. We posit that future studies will benefit from the inclusion of robophysical models to help understand the complexities of nest construction in nature.

\backmatter

\bmhead{Supplementary information}

A video demonstrating robot excavation capabilities, in addition to examples of success and failure modes, accompanies this article. Please see attached video file.

\bmhead{Acknowledgements}

The authors would like to thank various funding sources: L.T. was supported by NSF Grant \#NSF-IOS-2019799. D.S. was supported by NSF Physics of Living Systems (PoLS) and the Office of the Executive Vice President for Research (EVPR). J.H. was supported by the Georgia Tech President’s Undergraduate Research Award (PURA) program. 

\section*{Statements and Declarations}

\begin{itemize}
\item Competing interests: The authors declare no conflicts of interest or competing interests.
\end{itemize}

\noindent

\begin{appendices}
\section{Details of Robophysical Model Design and Sensing Strategy}\label{secA1}

\begin{figure}[!h]
    \centering
    \includegraphics[scale=0.65]{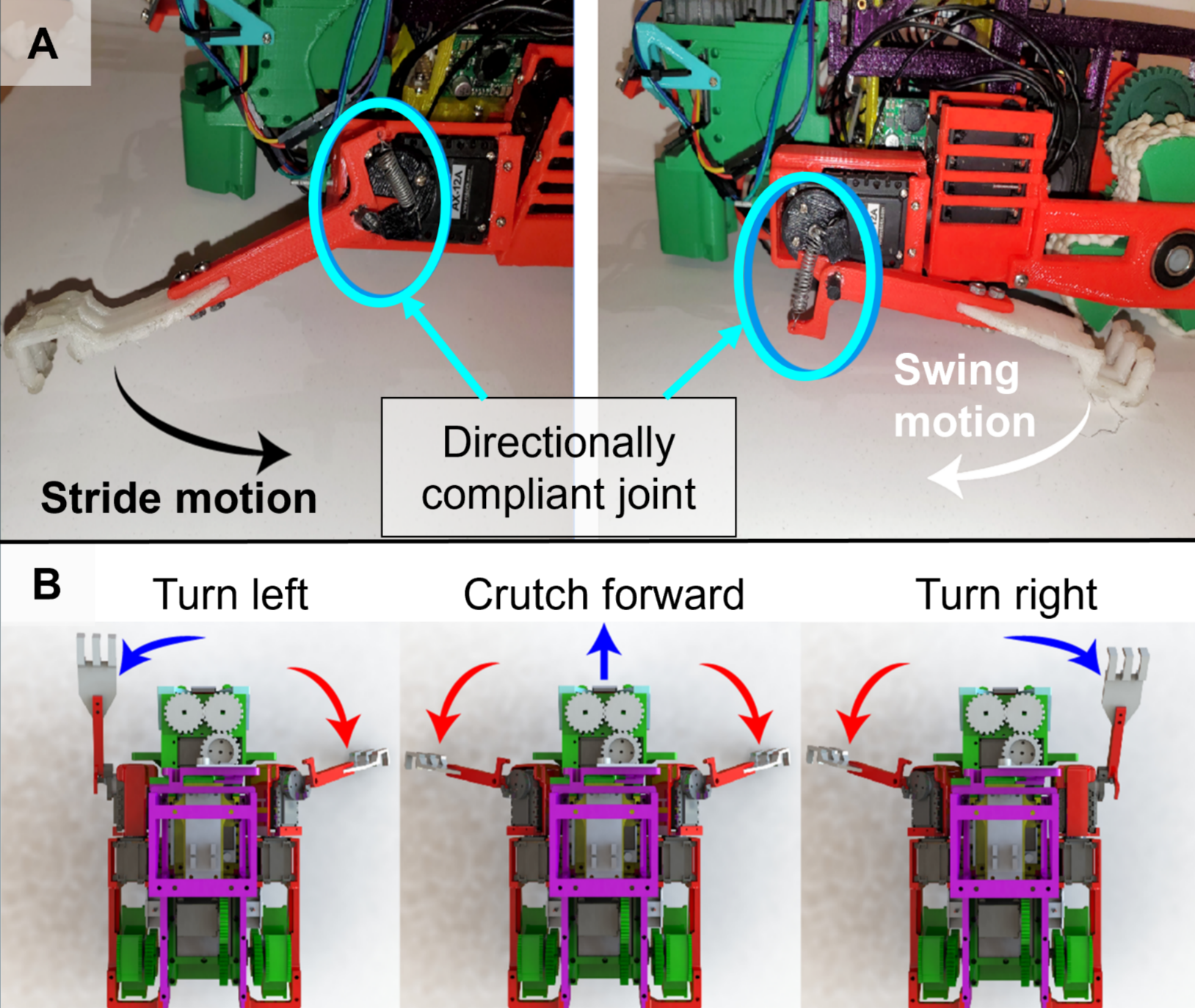}
    \caption{\textbf{Illustrations of robot locomotion strategies }\textbf{(A)} View of the spring-loaded directionally compliant joint attached to the claw assembly. Both the stride (protraction) and swing (retraction) phases are illustrated. \textbf{(B)} Schematic of the robot arm motion in red and resultant turning motion in blue. }
    \label{fig:robot_locomotion}
\end{figure}

The robot is equipped with a 5400 mAh 7.4 V LiPo battery, which directly powers the OpenCM9.04 microcontroller, and feeds into a Pololu Step-Up Voltage Regulator U3V70A set to 11.1 V for the motors. Under normal operation, to move forward or backward, the robot rotates its rear whegs. However, periodically, the system will instead use its limbs to propel itself forward in either a ``crutch" or ``sweep" maneuver. These maneuvers consist of a stride and swing phase (Figure \ref{fig:robot_locomotion}A) where the limb tip is in contact with the ground during the stride and is raised off the ground during swing. We use directionally compliant limbs, similar to that described in \cite{ozkan2020systematic}, to simplify the control needed for obstacle negotiation during the swing phase without impeding the thrust generated during the stride phase. 

During the ``crutch" motion, both limbs' roll motors are fixed to point outwards while the second motors rotate  each limb 180°, which constitutes a stride motion. This results in the robot moving forward while pitching upwards and eventually landing back on its chassis. This maneuver serves to free the robot from any potential obstacles encountered during material transport. During the ``sweep" motion, the motors on each limb are coordinated such that the stride length is maximized while the robot body maintains a fixed angle with respect to the ground. This maneuver provides less net propulsion than a crutch but it results in a less substantial impact with the ground at the end of the motion, making it preferable while carrying material. To turn, the robot uses these same maneuvers but with only a single active limb (Figure \ref{fig:robot_locomotion}B). By choosing between maneuvers, the robot is able to effectively locomote in heterogeneous environments such as those created during transport of cohesive material.

\subsubsection{Sensing Modalities}
The robot is equipped with a suite of sensors that receive environmental signals and guide subsequent behaviors. Both the Sparkfun ICM-20948 IMU and the Sparkfun VL6180 time-of-flight rangefinder are used to inform locomotion maneuvers. The rangefinder indicates when an obstacle is in front of the robot, which triggers a reversing motion. The IMU is used to indicate when body motion (specifically, pitching) is impeded, which in turn triggers a crutch maneuver. The Adafruit TCS34725 RGB color sensor enables the robot to react to fixed colored lights in its environment (such fixed environmental signals are used by other robotic swarm systems \cite{garnier2007alice, arvin2015cosvarphi}). Lastly, the robot is equipped with two systems to detect piles of previously deposited material that are independently operated by Dynamixel XL320 motors: the ``antenna" and the 2-megapixel ArduCAM. The antenna uses the motor encoder to detect the height of piles 10 cm in front of the robot and serves as a form of short-range sensing. The ArduCAM provides long-range sensing in the form of 320x240 monochrome images, which allows for basic vision methods that can detect piles over 10 cm in front of the robot. Overall, these sensors allow the agent to reliably locomote in obstacle laden scenarios and respond to environmental signals in the form of color or material deposits.

\section{Details of Robophysical Model Experimental Setup}\label{secA2}

\begin{figure}[!h]
    \centering
    \includegraphics[scale=0.5]{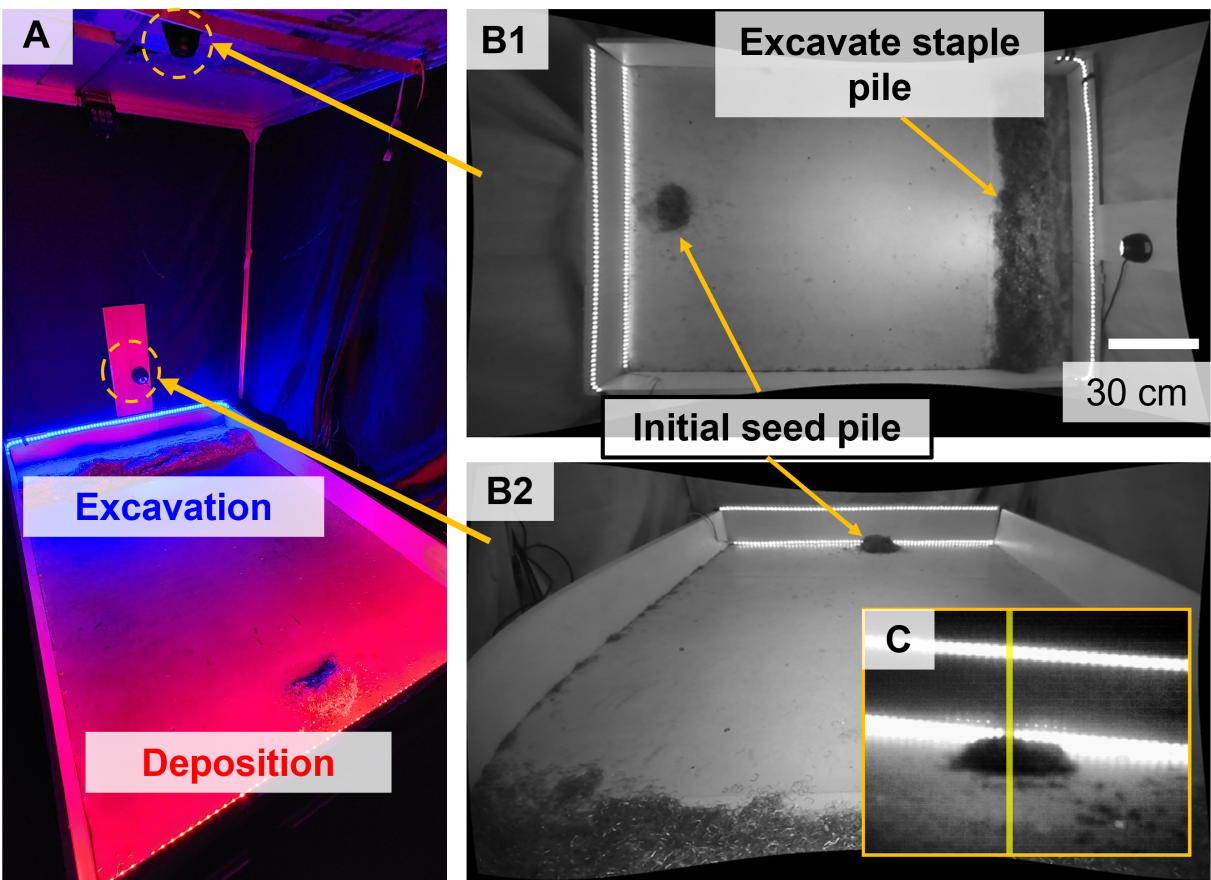}
    \caption{\textbf{(A)} Side view of the testing arena with colored lights. The excavation site, with a staple pile covering its base, is lit blue; the designated deposit wall is lit with red LEDs. The initial seed pile is situated along the deposit wall. \textbf{(B1-2)} Images from the ceiling and front night-vision camera. \textbf{(C)} Images taken by the ArduCAM which are used to detect existing piles.}
    \label{fig:arena}
\end{figure}

To explore strategies for manipulating and transporting cohesive material, we construct an arena (length = 1.8m, width = 1.2m) with colored lights serving as environmental signals to guide the robot (Figure \ref{fig:arena}A). The excavation and deposition areas are denoted with blue and red lights, respectively, and the arena is enclosed with blackout curtains to avoid outside interference. IR webcams provide an overhead view and a close-up view of the excavation area (Figure \ref{fig:arena}B1 and B2). Each trial begins with an initial seed pile that the agent detects. The robot then deposits material over the next two hours, after which the trial ends and we weigh the total amount of material transported. 

\section{Details of Robophysical Model Finite State Machine}\label{secA3}

\begin{figure}[!h]
    \centering
    \includegraphics[scale=0.65]{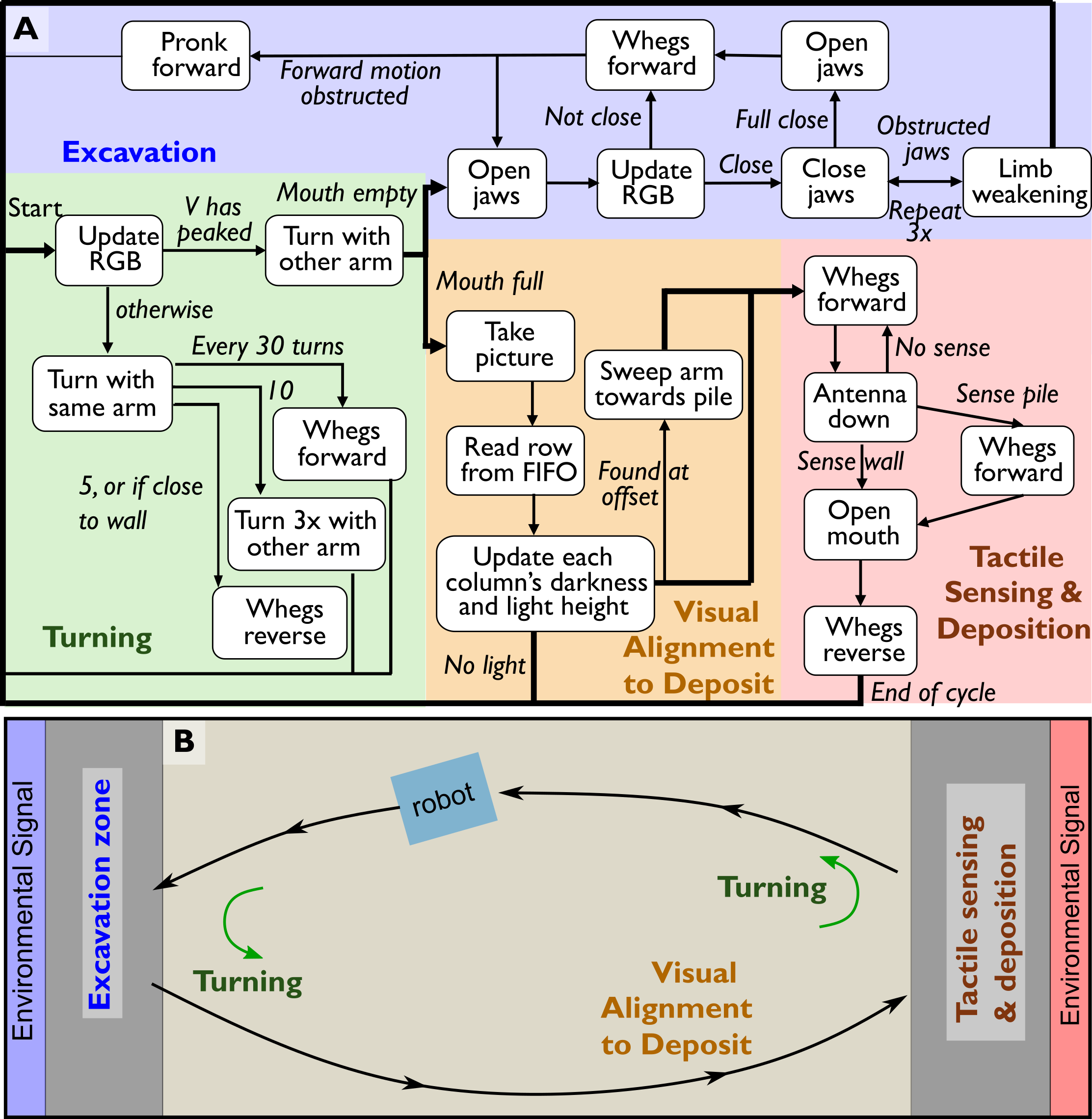}
    \caption{\textbf{(A)} State machine diagram for excavation and deposit sequences. Different stages of operation are indicated with different colors. Bold lines indicate a transition in state between any of these four stages. \textbf{(B)} Visual representation of various robot states within its testing arena.}
    \label{fig:state_machine}
\end{figure}

The robotic agent is equipped with a finite state machine (FSM) (Figure \ref{fig:state_machine}) based on the testing environment described above. The FSM embodies the different behaviors needed for the robotic agent to excavate material, transport it to the deposition site, locate an existing deposit, and finally return to the excavation site to repeat the process. The transitions between behaviors are governed by environmental signals which are detected via onboard sensors. Turning is initiated after successful excavation or deposition and is only stopped once the RGB sensor detects a local maximum in the desired color value. Once the robot is moving towards the excavation site (blue), it will begin the excavation procedure until it detects that material has remained within its jaws.

While the robot is moving to deposit, it takes a picture using the ArduCAM and searches for a pile by counting the number of ``dark" pixels (below a threshold) that exist in a column, for which two disconnected ``bright" regions (maximum intensity) are also present. The robot then chooses the largest group of connected columns within an image. This method amounts to looking for the two LED strips that define the excavation site (the bright regions) and searching for shadows (dark pixels) that correspond to piles of staples (Figure \ref{fig:arena}C). The robot then turns towards the largest group and transitions to the final state of the FSM. Here, it will search with its antenna and deposit its material either at the existing pile or at the wall and then begin turning towards the excavation site, thus completing one cycle of material excavation and deposition. 

\end{appendices}

\bibliography{staple-robot-construction} 


\begin{thebibliography}{45}
\ifx \bisbn   \undefined \def \bisbn  #1{ISBN #1}\fi
\ifx \binits  \undefined \def \binits#1{#1}\fi
\ifx \bauthor  \undefined \def \bauthor#1{#1}\fi
\ifx \batitle  \undefined \def \batitle#1{#1}\fi
\ifx \bjtitle  \undefined \def \bjtitle#1{#1}\fi
\ifx \bvolume  \undefined \def \bvolume#1{\textbf{#1}}\fi
\ifx \byear  \undefined \def \byear#1{#1}\fi
\ifx \bissue  \undefined \def \bissue#1{#1}\fi
\ifx \bfpage  \undefined \def \bfpage#1{#1}\fi
\ifx \blpage  \undefined \def \blpage #1{#1}\fi
\ifx \burl  \undefined \def \burl#1{\textsf{#1}}\fi
\ifx \doiurl  \undefined \def \doiurl#1{\url{https://doi.org/#1}}\fi
\ifx \betal  \undefined \def \betal{\textit{et al.}}\fi
\ifx \binstitute  \undefined \def \binstitute#1{#1}\fi
\ifx \binstitutionaled  \undefined \def \binstitutionaled#1{#1}\fi
\ifx \bctitle  \undefined \def \bctitle#1{#1}\fi
\ifx \beditor  \undefined \def \beditor#1{#1}\fi
\ifx \bpublisher  \undefined \def \bpublisher#1{#1}\fi
\ifx \bbtitle  \undefined \def \bbtitle#1{#1}\fi
\ifx \bedition  \undefined \def \bedition#1{#1}\fi
\ifx \bseriesno  \undefined \def \bseriesno#1{#1}\fi
\ifx \blocation  \undefined \def \blocation#1{#1}\fi
\ifx \bsertitle  \undefined \def \bsertitle#1{#1}\fi
\ifx \bsnm \undefined \def \bsnm#1{#1}\fi
\ifx \bsuffix \undefined \def \bsuffix#1{#1}\fi
\ifx \bparticle \undefined \def \bparticle#1{#1}\fi
\ifx \barticle \undefined \def \barticle#1{#1}\fi
\bibcommenthead
\ifx \bconfdate \undefined \def \bconfdate #1{#1}\fi
\ifx \botherref \undefined \def \botherref #1{#1}\fi
\ifx \url \undefined \def \url#1{\textsf{#1}}\fi
\ifx \bchapter \undefined \def \bchapter#1{#1}\fi
\ifx \bbook \undefined \def \bbook#1{#1}\fi
\ifx \bcomment \undefined \def \bcomment#1{#1}\fi
\ifx \oauthor \undefined \def \oauthor#1{#1}\fi
\ifx \citeauthoryear \undefined \def \citeauthoryear#1{#1}\fi
\ifx \endbibitem  \undefined \def \endbibitem {}\fi
\ifx \bconflocation  \undefined \def \bconflocation#1{#1}\fi
\ifx \arxivurl  \undefined \def \arxivurl#1{\textsf{#1}}\fi
\csname PreBibitemsHook\endcsname

\bibitem[\protect\citeauthoryear{Petersen et~al.}{2019}]{petersen_review_2019}
\begin{barticle}
\bauthor{\bsnm{Petersen}, \binits{K.H.}},
\bauthor{\bsnm{Napp}, \binits{N.}},
\bauthor{\bsnm{Stuart-Smith}, \binits{R.}},
\bauthor{\bsnm{Rus}, \binits{D.}},
\bauthor{\bsnm{Kovac}, \binits{M.}}:
\batitle{A review of collective robotic construction}.
\bjtitle{Science Robotics}
\bvolume{4}(\bissue{28}),
\bfpage{8479}
(\byear{2019})
\doiurl{10.1126/scirobotics.aau8479}
\end{barticle}
\endbibitem

\bibitem[\protect\citeauthoryear{Werfel et~al.}{2014}]{werfel_designing_2014}
\begin{barticle}
\bauthor{\bsnm{Werfel}, \binits{J.}},
\bauthor{\bsnm{Petersen}, \binits{K.}},
\bauthor{\bsnm{Nagpal}, \binits{R.}}:
\batitle{Designing {Collective} {Behavior} in a {Termite}-{Inspired} {Robot} {Construction} {Team}}.
\bjtitle{Science}
\bvolume{343}(\bissue{6172}),
\bfpage{754}--\blpage{758}
(\byear{2014})
\doiurl{10.1126/science.1245842}
\end{barticle}
\endbibitem

\bibitem[\protect\citeauthoryear{Seo et~al.}{2013}]{seo_assembly_2013}
\begin{bchapter}
\bauthor{\bsnm{Seo}, \binits{J.}},
\bauthor{\bsnm{Yim}, \binits{M.}},
\bauthor{\bsnm{Kumar}, \binits{V.}}:
\bctitle{Assembly planning for planar structures of a brick wall pattern with rectangular modular robots}.
In: \bbtitle{2013 IEEE International Conference on Automation Science and Engineering (CASE)},
pp. \bfpage{1016}--\blpage{1021}
(\byear{2013}).
\doiurl{10.1109/CoASE.2013.6653996}
\end{bchapter}
\endbibitem

\bibitem[\protect\citeauthoryear{Napp and Nagpal}{2014}]{napp_distributed_2014}
\begin{bchapter}
\bauthor{\bsnm{Napp}, \binits{N.}},
\bauthor{\bsnm{Nagpal}, \binits{R.}}:
\bctitle{Distributed {Amorphous} {Ramp} {Construction} in {Unstructured} {Environments}}.
In: \bbtitle{Distributed {Autonomous} {Robotic} {Systems}}.
\bsertitle{Springer {Tracts} in {Advanced} {Robotics}},
pp. \bfpage{105}--\blpage{119}
(\byear{2014}).
\doiurl{10.1007/978-3-642-55146-8_8}
\end{bchapter}
\endbibitem

\bibitem[\protect\citeauthoryear{Saboia et~al.}{2019}]{saboia_autonomous_2019}
\begin{barticle}
\bauthor{\bsnm{Saboia}, \binits{M.}},
\bauthor{\bsnm{Thangavelu}, \binits{V.}},
\bauthor{\bsnm{Napp}, \binits{N.}}:
\batitle{Autonomous multi-material construction with a heterogeneous robot team}.
\bjtitle{Robotics and Autonomous Systems}
\bvolume{121},
\bfpage{103239}
(\byear{2019})
\doiurl{10.1016/j.robot.2019.07.009}
\end{barticle}
\endbibitem

\bibitem[\protect\citeauthoryear{Napp et~al.}{2012}]{napp_materials_2012}
\begin{bchapter}
\bauthor{\bsnm{Napp}, \binits{N.}},
\bauthor{\bsnm{Rappoli}, \binits{O.R.}},
\bauthor{\bsnm{Wu}, \binits{J.M.}},
\bauthor{\bsnm{Nagpal}, \binits{R.}}:
\bctitle{Materials and mechanisms for amorphous robotic construction}.
In: \bbtitle{2012 {IEEE}/{RSJ} {International} {Conference} on {Intelligent} {Robots} and {Systems}},
pp. \bfpage{4879}--\blpage{4885}
(\byear{2012}).
\doiurl{10.1109/IROS.2012.6385718}
\end{bchapter}
\endbibitem

\bibitem[\protect\citeauthoryear{Saboia~da Silva}{2019}]{saboia_da_silva_autonomous_2019}
\begin{botherref}
\oauthor{\bsnm{Silva}, \binits{M.}}:
Autonomous {Adaptive} {Modification} of {Unstructured} {Environments}.
Ph.{D}. thesis,
State University of New York at Buffalo
(2019).
\url{https://www.proquest.com/docview/2320957977/abstract/4A8BA4DC04954E39PQ/1}
\end{botherref}
\endbibitem

\bibitem[\protect\citeauthoryear{Furrer et~al.}{2017}]{furrer_autonomous_2017}
\begin{bchapter}
\bauthor{\bsnm{Furrer}, \binits{F.}},
\bauthor{\bsnm{Wermelinger}, \binits{M.}},
\bauthor{\bsnm{Yoshida}, \binits{H.}},
\bauthor{\bsnm{Gramazio}, \binits{F.}},
\bauthor{\bsnm{Kohler}, \binits{M.}},
\bauthor{\bsnm{Siegwart}, \binits{R.}},
\bauthor{\bsnm{Hutter}, \binits{M.}}:
\bctitle{Autonomous robotic stone stacking with online next best object target pose planning}.
In: \bbtitle{2017 {IEEE} {International} {Conference} on {Robotics} and {Automation} ({ICRA})},
pp. \bfpage{2350}--\blpage{2356}
(\byear{2017}).
\doiurl{10.1109/ICRA.2017.7989272}
\end{bchapter}
\endbibitem

\bibitem[\protect\citeauthoryear{Thangavelu et~al.}{2018}]{thangavelu_dry_2018}
\begin{bchapter}
\bauthor{\bsnm{Thangavelu}, \binits{V.}},
\bauthor{\bsnm{Liu}, \binits{Y.}},
\bauthor{\bsnm{Saboia}, \binits{M.}},
\bauthor{\bsnm{Napp}, \binits{N.}}:
\bctitle{Dry {Stacking} for {Automated} {Construction} with {Irregular} {Objects}}.
In: \bbtitle{2018 {IEEE} {International} {Conference} on {Robotics} and {Automation} ({ICRA})},
pp. \bfpage{4782}--\blpage{4789}
(\byear{2018}).
\doiurl{10.1109/ICRA.2018.8460562}
\end{bchapter}
\endbibitem

\bibitem[\protect\citeauthoryear{Matl}{2021}]{matl_interactive_2021}
\begin{botherref}
\oauthor{\bsnm{Matl}, \binits{C.C.}}:
Interactive perception for robotic manipulation of liquids, grains, and doughs.
Ph{D} {T}hesis,
University of California, Berkeley
(2021)
\end{botherref}
\endbibitem

\bibitem[\protect\citeauthoryear{Schenck et~al.}{2017}]{pmlr-v78-schenck17a}
\begin{bchapter}
\bauthor{\bsnm{Schenck}, \binits{C.}},
\bauthor{\bsnm{Tompson}, \binits{J.}},
\bauthor{\bsnm{Levine}, \binits{S.}},
\bauthor{\bsnm{Fox}, \binits{D.}}:
\bctitle{Learning robotic manipulation of granular media}.
In: \bbtitle{Proceedings of the 1st Annual Conference on Robot Learning}.
\bsertitle{Proceedings of Machine Learning Research},
vol. \bseriesno{78},
pp. \bfpage{239}--\blpage{248}
(\byear{2017})
\end{bchapter}
\endbibitem

\bibitem[\protect\citeauthoryear{Tuomainen et~al.}{2022}]{tuomainen_manipulation_2022}
\begin{barticle}
\bauthor{\bsnm{Tuomainen}, \binits{N.}},
\bauthor{\bsnm{Blanco-Mulero}, \binits{D.}},
\bauthor{\bsnm{Kyrki}, \binits{V.}}:
\batitle{Manipulation of granular materials by learning particle interactions}.
\bjtitle{{IEEE} Robotics and Automation Letters}
\bvolume{7}(\bissue{2}),
\bfpage{5663}--\blpage{5670}
(\byear{2022})
\doiurl{10.1109/LRA.2022.3158382}
\end{barticle}
\endbibitem

\bibitem[\protect\citeauthoryear{Keller and Jaeger}{2016}]{keller2016aleatory}
\begin{barticle}
\bauthor{\bsnm{Keller}, \binits{S.}},
\bauthor{\bsnm{Jaeger}, \binits{H.M.}}:
\batitle{Aleatory architectures}.
\bjtitle{Granular Matter}
\bvolume{18},
\bfpage{1}--\blpage{11}
(\byear{2016})
\doiurl{10.1007/s10035-016-0629-x}
\end{barticle}
\endbibitem

\bibitem[\protect\citeauthoryear{Murphy et~al.}{2017}]{murphy2017aleatory}
\begin{barticle}
\bauthor{\bsnm{Murphy}, \binits{K.}},
\bauthor{\bsnm{Roth}, \binits{L.}},
\bauthor{\bsnm{Peterman}, \binits{D.}},
\bauthor{\bsnm{Jaeger}, \binits{H.}}:
\batitle{Aleatory construction based on jamming: Stability through self-confinement}.
\bjtitle{Architectural Design}
\bvolume{87}(\bissue{4}),
\bfpage{74}--\blpage{81}
(\byear{2017})
\doiurl{10.1002/ad.2198}
\end{barticle}
\endbibitem

\bibitem[\protect\citeauthoryear{Dierichs and Menges}{2016}]{dierichs_towards_2016}
\begin{barticle}
\bauthor{\bsnm{Dierichs}, \binits{K.}},
\bauthor{\bsnm{Menges}, \binits{A.}}:
\batitle{Towards an aggregate architecture: designed granular systems as programmable matter in architecture}.
\bjtitle{Granular Matter}
\bvolume{18}(\bissue{2}),
\bfpage{25}
(\byear{2016})
\doiurl{10.1007/s10035-016-0631-3}
\end{barticle}
\endbibitem

\bibitem[\protect\citeauthoryear{Dierichs and Menges}{2021}]{dierichs2021designing}
\begin{barticle}
\bauthor{\bsnm{Dierichs}, \binits{K.}},
\bauthor{\bsnm{Menges}, \binits{A.}}:
\batitle{Designing architectural materials: from granular form to functional granular material}.
\bjtitle{Bioinspiration \& Biomimetics}
\bvolume{16}(\bissue{6}),
\bfpage{065010}
(\byear{2021})
\doiurl{10.1088/1748-3190/ac2987}
\end{barticle}
\endbibitem

\bibitem[\protect\citeauthoryear{Zhao et~al.}{2016}]{zhao2016packings}
\begin{barticle}
\bauthor{\bsnm{Zhao}, \binits{Y.}},
\bauthor{\bsnm{Liu}, \binits{K.}},
\bauthor{\bsnm{Zheng}, \binits{M.}},
\bauthor{\bsnm{Bar{\'e}s}, \binits{J.}},
\bauthor{\bsnm{Dierichs}, \binits{K.}},
\bauthor{\bsnm{Menges}, \binits{A.}},
\bauthor{\bsnm{Behringer}, \binits{R.P.}}:
\batitle{Packings of 3d stars: stability and structure}.
\bjtitle{Granular Matter}
\bvolume{18}(\bissue{2}),
\bfpage{24}
(\byear{2016})
\doiurl{10.1007/s10035-016-0606-4}
\end{barticle}
\endbibitem

\bibitem[\protect\citeauthoryear{Savoie et~al.}{2023}]{savoie_amorphous_2023}
\begin{barticle}
\bauthor{\bsnm{Savoie}, \binits{W.}},
\bauthor{\bsnm{Tuazon}, \binits{H.}},
\bauthor{\bsnm{Tiwari}, \binits{I.}},
\bauthor{\bsnm{Bhamla}, \binits{S.M.}},
\bauthor{\bsnm{Goldman}, \binits{D.I.}}:
\batitle{Amorphous entangled active matter}.
\bjtitle{Soft Matter}
\bvolume{19}(\bissue{10}),
\bfpage{1952}--\blpage{1965}
(\byear{2023})
\doiurl{10.1039/D2SM01573K}
\end{barticle}
\endbibitem

\bibitem[\protect\citeauthoryear{Savoie et~al.}{2019}]{savoie_smarticles_2019}
\begin{botherref}
\oauthor{\bsnm{Savoie}, \binits{W.}},
\oauthor{\bsnm{Berrueta}, \binits{T.}},
\oauthor{\bsnm{Jackson}, \binits{Z.}},
\oauthor{\bsnm{Pervan}, \binits{A.}},
\oauthor{\bsnm{Warkentin}, \binits{R.}},
\oauthor{\bsnm{Li}, \binits{S.}},
\oauthor{\bsnm{Murphey}, \binits{T.D.}},
\oauthor{\bsnm{Wiesenfeld}, \binits{K.}},
\oauthor{\bsnm{Goldman}, \binits{D.I.}}:
A robot made of robots: Emergent transport and control of a smarticle ensemble.
Science
\textbf{4}(34)
(2019)
\doiurl{10.1126/scirobotics.aax4316}
\end{botherref}
\endbibitem

\bibitem[\protect\citeauthoryear{Gravish et~al.}{2012}]{gravish_entangled_2012}
\begin{barticle}
\bauthor{\bsnm{Gravish}, \binits{N.}},
\bauthor{\bsnm{Franklin}, \binits{S.V.}},
\bauthor{\bsnm{Hu}, \binits{D.L.}},
\bauthor{\bsnm{Goldman}, \binits{D.I.}}:
\batitle{Entangled {Granular} {Media}}.
\bjtitle{Physical Review Letters}
\bvolume{108}(\bissue{20}),
\bfpage{208001}
(\byear{2012})
\doiurl{10.1103/PhysRevLett.108.208001}
\end{barticle}
\endbibitem

\bibitem[\protect\citeauthoryear{Gravish and I.~Goldman}{2016}]{gravish_entangled_2016}
\begin{bchapter}
\bauthor{\bsnm{Gravish}, \binits{N.}},
\bauthor{\bsnm{I.~Goldman}, \binits{D.}}:
\bctitle{Entangled {Granular} {Media}}.
In: \bbtitle{Fluids, {Colloids} and {Soft} {Materials}},
pp. \bfpage{341}--\blpage{354}
(\byear{2016}).
\doiurl{10.1002/9781119220510.ch17}
\end{bchapter}
\endbibitem

\bibitem[\protect\citeauthoryear{Franklin}{2014}]{franklin_extensional_2014}
\begin{barticle}
\bauthor{\bsnm{Franklin}, \binits{S.}}:
\batitle{Extensional rheology of entangled granular materials}.
\bjtitle{Europhysics Letters}
\bvolume{106}(\bissue{5}),
\bfpage{58004}
(\byear{2014})
\doiurl{10.1209/0295-5075/106/58004}
\end{barticle}
\endbibitem

\bibitem[\protect\citeauthoryear{Sohn et~al.}{2025}]{pezeshki_tunable_2024}
\begin{barticle}
\bauthor{\bsnm{Sohn}, \binits{Y.}},
\bauthor{\bsnm{Pezeshki}, \binits{S.}},
\bauthor{\bsnm{Barthelat}, \binits{F.}}:
\batitle{Tuning geometry in staple-like entangled particles: ``pick-up” experiments and monte carlo simulations}.
\bjtitle{Granular Matter}
\bvolume{27},
\bfpage{55}
(\byear{2025})
\doiurl{10.1007/s10035-025-01531-w}
\end{barticle}
\endbibitem

\bibitem[\protect\citeauthoryear{Karapiperis et~al.}{2022}]{karapiperis_stress_2022}
\begin{barticle}
\bauthor{\bsnm{Karapiperis}, \binits{K.}},
\bauthor{\bsnm{Monfared}, \binits{S.}},
\bauthor{\bsnm{Macedo}, \binits{R.B.d.}},
\bauthor{\bsnm{Richardson}, \binits{S.}},
\bauthor{\bsnm{Andrade}, \binits{J.E.}}:
\batitle{Stress transmission in entangled granular structures}.
\bjtitle{Granular Matter}
\bvolume{24}(\bissue{3}),
\bfpage{91}
(\byear{2022})
\doiurl{10.1007/s10035-022-01252-4}
\end{barticle}
\endbibitem

\bibitem[\protect\citeauthoryear{Marschall et~al.}{2015}]{marschall_compression-_2015}
\begin{barticle}
\bauthor{\bsnm{Marschall}, \binits{T.A.}},
\bauthor{\bsnm{Franklin}, \binits{S.V.}},
\bauthor{\bsnm{Teitel}, \binits{S.}}:
\batitle{Compression- and shear-driven jamming of {U}-shaped particles in two dimensions}.
\bjtitle{Granular Matter}
\bvolume{17}(\bissue{1}),
\bfpage{121}--\blpage{133}
(\byear{2015})
\doiurl{10.1007/s10035-014-0540-2}
\end{barticle}
\endbibitem

\bibitem[\protect\citeauthoryear{Aguilar et~al.}{2016}]{aguilar2016review}
\begin{barticle}
\bauthor{\bsnm{Aguilar}, \binits{J.}},
\bauthor{\bsnm{Zhang}, \binits{T.}},
\bauthor{\bsnm{Qian}, \binits{F.}},
\bauthor{\bsnm{Kingsbury}, \binits{M.}},
\bauthor{\bsnm{McInroe}, \binits{B.}},
\bauthor{\bsnm{Mazouchova}, \binits{N.}},
\bauthor{\bsnm{Li}, \binits{C.}},
\bauthor{\bsnm{Maladen}, \binits{R.}},
\bauthor{\bsnm{Gong}, \binits{C.}},
\bauthor{\bsnm{Travers}, \binits{M.}}, \betal:
\batitle{A review on locomotion robophysics: the study of movement at the intersection of robotics, soft matter and dynamical systems}.
\bjtitle{Reports on Progress in Physics}
\bvolume{79}(\bissue{11}),
\bfpage{110001}
(\byear{2016})
\doiurl{10.1088/0034-4885/79/11/110001}
\end{barticle}
\endbibitem

\bibitem[\protect\citeauthoryear{Quinn et~al.}{2002}]{quinn2002improved}
\begin{bchapter}
\bauthor{\bsnm{Quinn}, \binits{R.D.}},
\bauthor{\bsnm{Offi}, \binits{J.T.}},
\bauthor{\bsnm{Kingsley}, \binits{D.A.}},
\bauthor{\bsnm{Ritzmann}, \binits{R.E.}}:
\bctitle{Improved mobility through abstracted biological principles}.
In: \bbtitle{IEEE/RSJ International Conference on Intelligent Robots and Systems},
vol. \bseriesno{3},
pp. \bfpage{2652}--\blpage{2657}
(\byear{2002}).
\doiurl{10.1109/IRDS.2002.1041670}
\end{bchapter}
\endbibitem

\bibitem[\protect\citeauthoryear{von Ehrenfried}{2020}]{von_ehrenfried_artemis_2020}
\begin{bchapter}
\bauthor{\bsnm{Ehrenfried}, \binits{M.}}:
\bctitle{The {Artemis} {Lunar} {Program} {Overview}}.
In: \bbtitle{The {Artemis} {Lunar} {Program}: {Returning} {People} To the {Moon}},
pp. \bfpage{7}--\blpage{47}
(\byear{2020}).
\doiurl{10.1007/978-3-030-38513-2_2}
\end{bchapter}
\endbibitem

\bibitem[\protect\citeauthoryear{Khoshnevis}{2004}]{khoshnevis_automated_2004}
\begin{barticle}
\bauthor{\bsnm{Khoshnevis}, \binits{B.}}:
\batitle{Automated construction by contour crafting—related robotics and information technologies}.
\bjtitle{Automation in Construction}
\bvolume{13}(\bissue{1}),
\bfpage{5}--\blpage{19}
(\byear{2004})
\doiurl{10.1016/j.autcon.2003.08.012}
\end{barticle}
\endbibitem

\bibitem[\protect\citeauthoryear{Rossi et~al.}{2024}]{rossi_computational_2024}
\begin{barticle}
\bauthor{\bsnm{Rossi}, \binits{M.}},
\bauthor{\bsnm{Sumini}, \binits{V.}},
\bauthor{\bsnm{Zanelli}, \binits{A.}},
\bauthor{\bsnm{Viscuso}, \binits{S.}}:
\batitle{Computational {Design} of an {Extreme} {Livable} {Lightweight} {Environment} on {Mars}}.
\bjtitle{Journal of Architectural Engineering}
\bvolume{30}(\bissue{2}),
\bfpage{04024009}
(\byear{2024})
\doiurl{10.1061/JAEIED.AEENG-1632}
\end{barticle}
\endbibitem

\bibitem[\protect\citeauthoryear{Mueller}{2017}]{mueller_construction_2017}
\begin{bchapter}
\bauthor{\bsnm{Mueller}, \binits{R.P.}}:
\bctitle{Construction with regolith}.
In: \bbtitle{The Technology and Future of In-Situ Resource Utilization (ISRU): A Capstone Graduate Seminar}
(\byear{2017})
\end{bchapter}
\endbibitem

\bibitem[\protect\citeauthoryear{Gerling and Von~Mammen}{2016}]{gerling_robotics_2016}
\begin{bchapter}
\bauthor{\bsnm{Gerling}, \binits{V.}},
\bauthor{\bsnm{Von~Mammen}, \binits{S.}}:
\bctitle{Robotics for {Self}-{Organised} {Construction}}.
In: \bbtitle{2016 {IEEE} 1st {International} {Workshops} on {Foundations} and {Applications} of {Self}* {Systems} ({FAS}*{W})},
pp. \bfpage{162}--\blpage{167}
(\byear{2016}).
\doiurl{10.1109/FAS-W.2016.45}
\end{bchapter}
\endbibitem

\bibitem[\protect\citeauthoryear{Pradhananga et~al.}{2021}]{pradhananga_identifying_2021}
\begin{barticle}
\bauthor{\bsnm{Pradhananga}, \binits{P.}},
\bauthor{\bsnm{ElZomor}, \binits{M.}},
\bauthor{\bsnm{Santi~Kasabdji}, \binits{G.}}:
\batitle{Identifying the {Challenges} to {Adopting} {Robotics} in the {US} {Construction} {Industry}}.
\bjtitle{Journal of Construction Engineering and Management}
\bvolume{147}(\bissue{5}),
\bfpage{05021003}
(\byear{2021})
\doiurl{10.1061/(ASCE)CO.1943-7862.0002007}
\end{barticle}
\endbibitem

\bibitem[\protect\citeauthoryear{Gregory et~al.}{2016}]{gregory_application_2016}
\begin{bchapter}
\bauthor{\bsnm{Gregory}, \binits{J.}},
\bauthor{\bsnm{Fink}, \binits{J.}},
\bauthor{\bsnm{Stump}, \binits{E.}},
\bauthor{\bsnm{Twigg}, \binits{J.}},
\bauthor{\bsnm{Rogers}, \binits{J.}},
\bauthor{\bsnm{Baran}, \binits{D.}},
\bauthor{\bsnm{Fung}, \binits{N.}},
\bauthor{\bsnm{Young}, \binits{S.}}:
\bctitle{Application of {Multi}-{Robot} {Systems} to {Disaster}-{Relief} {Scenarios} with {Limited} {Communication}}.
In: \bbtitle{Field and {Service} {Robotics}: {Results} of the 10th {International} {Conference}},
pp. \bfpage{639}--\blpage{653}
(\byear{2016}).
\doiurl{10.1007/978-3-319-27702-8_42} .
\burl{https://doi.org/10.1007/978-3-319-27702-8_42}
\end{bchapter}
\endbibitem

\bibitem[\protect\citeauthoryear{Hansell}{2005}]{hansell_animal_2005}
\begin{botherref}
\oauthor{\bsnm{Hansell}, \binits{M.H.}}:
Animal {Architecture}, {Oxford} {University} {Press}
(2005)
\end{botherref}
\endbibitem

\bibitem[\protect\citeauthoryear{Tschinkel}{2013}]{tschinkel_fire_2013}
\begin{botherref}
\oauthor{\bsnm{Tschinkel}, \binits{W.R.}}:
The {Fire} {Ants}. {Harvard} {University} {Press}
(2013)
\end{botherref}
\endbibitem

\bibitem[\protect\citeauthoryear{Hölldobler and Wilson}{1990}]{holldobler_ants_1990}
\begin{botherref}
\oauthor{\bsnm{Hölldobler}, \binits{B.}},
\oauthor{\bsnm{Wilson}, \binits{E.O.}}:
The {Ants}. {Harvard} {University} {Press}
(1990)
\end{botherref}
\endbibitem

\bibitem[\protect\citeauthoryear{Monaenkova et~al.}{2015}]{monaenkova_behavioral_2015}
\begin{barticle}
\bauthor{\bsnm{Monaenkova}, \binits{D.}},
\bauthor{\bsnm{Gravish}, \binits{N.}},
\bauthor{\bsnm{Rodriguez}, \binits{G.}},
\bauthor{\bsnm{Kutner}, \binits{R.}},
\bauthor{\bsnm{Goodisman}, \binits{M.A.D.}},
\bauthor{\bsnm{Goldman}, \binits{D.I.}}:
\batitle{Behavioral and mechanical determinants of collective subsurface nest excavation}.
\bjtitle{Journal of Experimental Biology}
\bvolume{218}(\bissue{9}),
\bfpage{1295}--\blpage{1305}
(\byear{2015})
\doiurl{10.1242/jeb.113795}
\end{barticle}
\endbibitem

\bibitem[\protect\citeauthoryear{Larsen et~al.}{2021}]{larsen_dam_2021}
\begin{barticle}
\bauthor{\bsnm{Larsen}, \binits{A.}},
\bauthor{\bsnm{Larsen}, \binits{J.R.}},
\bauthor{\bsnm{Lane}, \binits{S.N.}}:
\batitle{Dam builders and their works: {Beaver} influences on the structure and function of river corridor hydrology, geomorphology, biogeochemistry and ecosystems}.
\bjtitle{Earth-Science Reviews}
\bvolume{218},
\bfpage{103623}
(\byear{2021})
\doiurl{10.1016/j.earscirev.2021.103623}
\end{barticle}
\endbibitem

\bibitem[\protect\citeauthoryear{Andrade-Silva et~al.}{2021}]{andrade-silva_cohesion_2021}
\begin{barticle}
\bauthor{\bsnm{Andrade-Silva}, \binits{I.}},
\bauthor{\bsnm{Godefroy}, \binits{T.}},
\bauthor{\bsnm{Pouliquen}, \binits{O.}},
\bauthor{\bsnm{Marthelot}, \binits{J.}}:
\batitle{Cohesion of bird nests}.
\bjtitle{EPJ Web of Conferences}
\bvolume{249},
\bfpage{06014}
(\byear{2021})
\doiurl{10.1051/epjconf/202124906014}
\end{barticle}
\endbibitem

\bibitem[\protect\citeauthoryear{Weiner et~al.}{2020}]{weiner_mechanics_2020}
\begin{barticle}
\bauthor{\bsnm{Weiner}, \binits{N.}},
\bauthor{\bsnm{Bhosale}, \binits{Y.}},
\bauthor{\bsnm{Gazzola}, \binits{M.}},
\bauthor{\bsnm{King}, \binits{H.}}:
\batitle{Mechanics of randomly packed filaments—{The} “bird nest” as meta-material}.
\bjtitle{Journal of Applied Physics}
\bvolume{127}(\bissue{5}),
\bfpage{050902}
(\byear{2020})
\doiurl{10.1063/1.5132809}
\end{barticle}
\endbibitem

\bibitem[\protect\citeauthoryear{Bhosale et~al.}{2022}]{bhosale_micromechanical_2022}
\begin{barticle}
\bauthor{\bsnm{Bhosale}, \binits{Y.}},
\bauthor{\bsnm{Weiner}, \binits{N.}},
\bauthor{\bsnm{Butler}, \binits{A.}},
\bauthor{\bsnm{Kim}, \binits{S.H.}},
\bauthor{\bsnm{Gazzola}, \binits{M.}},
\bauthor{\bsnm{King}, \binits{H.}}:
\batitle{Micromechanical {Origin} of {Plasticity} and {Hysteresis} in {Nestlike} {Packings}}.
\bjtitle{Physical Review Letters}
\bvolume{128}(\bissue{19}),
\bfpage{198003}
(\byear{2022})
\doiurl{10.1103/PhysRevLett.128.198003}
\end{barticle}
\endbibitem

\bibitem[\protect\citeauthoryear{Ozkan-Aydin et~al.}{2020}]{ozkan2020systematic}
\begin{bchapter}
\bauthor{\bsnm{Ozkan-Aydin}, \binits{Y.}},
\bauthor{\bsnm{Chong}, \binits{B.}},
\bauthor{\bsnm{Aydin}, \binits{E.}},
\bauthor{\bsnm{Goldman}, \binits{D.I.}}:
\bctitle{A systematic approach to creating terrain-capable hybrid soft/hard myriapod robots}.
In: \bbtitle{2020 3rd IEEE International Conference on Soft Robotics (RoboSoft)},
pp. \bfpage{156}--\blpage{163}
(\byear{2020}).
\doiurl{10.1109/RoboSoft48309.2020.9116022}
\end{bchapter}
\endbibitem

\bibitem[\protect\citeauthoryear{Garnier et~al.}{2007}]{garnier2007alice}
\begin{bchapter}
\bauthor{\bsnm{Garnier}, \binits{S.}},
\bauthor{\bsnm{Tache}, \binits{F.}},
\bauthor{\bsnm{Combe}, \binits{M.}},
\bauthor{\bsnm{Grimal}, \binits{A.}},
\bauthor{\bsnm{Theraulaz}, \binits{G.}}:
\bctitle{Alice in pheromone land: An experimental setup for the study of ant-like robots}.
In: \bbtitle{2007 IEEE Swarm Intelligence Symposium},
pp. \bfpage{37}--\blpage{44}
(\byear{2007}).
\doiurl{10.1109/SIS.2007.368024}
\end{bchapter}
\endbibitem

\bibitem[\protect\citeauthoryear{Arvin et~al.}{2015}]{arvin2015cosvarphi}
\begin{bchapter}
\bauthor{\bsnm{Arvin}, \binits{F.}},
\bauthor{\bsnm{Krajn{\'\i}k}, \binits{T.}},
\bauthor{\bsnm{Turgut}, \binits{A.E.}},
\bauthor{\bsnm{Yue}, \binits{S.}}:
\bctitle{Cos$\phi$: Artificial pheromone system for robotic swarms research}.
In: \bbtitle{2015 IEEE/RSJ International Conference on Intelligent Robots and Systems (IROS)},
pp. \bfpage{407}--\blpage{412}
(\byear{2015}).
\doiurl{10.1109/IROS.2015.7353405}
\end{bchapter}
\endbibitem

\end{thebibliography}

\end{document}